\pdfoutput=1

\documentclass[11pt,a4paper]{article}
\usepackage[hyperref]{acl}
\usepackage{times}
\usepackage{inconsolata}
\usepackage{latexsym}
\usepackage{danudefs}
\usepackage{algorithmic}
\usepackage{algorithm}
\usepackage{multirow}
\usepackage{multicol}
\usepackage{color}
\usepackage{booktabs}
\usepackage{xspace}
\usepackage{url}
\usepackage{braket}
\usepackage[T1]{fontenc}
\usepackage[utf8]{inputenc}
\usepackage{graphicx}
\usepackage{microtype}
\usepackage{pifont}
\usepackage{caption}
\usepackage{setspace}
\usepackage{enumitem}
\usepackage{tabularx}
\newcommand{\xmark}{\ding{55}}
\newcommand{\cmark}{\ding{52}} 

\usepackage[english]{babel}
\usepackage{hyperref}
\addto\captionsenglish{%
}
\addto\extrasenglish{%
}

\usepackage{acronym}

\usepackage{etoolbox}
\makeatletter
\newif\if@in@acrolist
\AtBeginEnvironment{acronym}{\@in@acrolisttrue}
\newrobustcmd{\LU}[2]{\if@in@acrolist#1\else#2\fi}

\newcommand{\ACF}[1]{{\@in@acrolisttrue\acf{#1}}}
\makeatother

\acrodef{MLM}[MLM]{Masked Language Model}
\acrodef{LM}[LM]{Language Model}
\acrodefplural{MLM}[MLMs]{\LU{M}{m}asked \LU{L}{l}anguage \LU{M}{m}odels}
\acrodef{LLM}[LLM]{Large Language Model}
\acrodefplural{LLM}[LLMs]{Large Language Models}
\acrodef{SoTA}[SoTA]{state-of-the-art}
\acrodef{ICL}{\LU{I}{i}n-cotext \LU{L}{l}earning}
\acrodef{SCD}{\LU{S}{s}emantic \LU{C}{c}hange \LU{D}{d}etection}
\acrodef{WiC}{Word-in-Context}
\acrodef{ITML}[ITML]{Information-Theoretic Metric Learning}
\acrodef{SDML}[SDML]{Semantic Distance Metric Learning}
\acrodef{GCR}[GCR]{Generative Commonsense Reasoning}
\acrodef{NLG}[NLG]{Natural Language Generation}
\acrodef{NLP}[NLP]{Natural Language Processing}
\acrodef{KG}[KG]{knowledge graph}
\acrodef{SFT}[SFT]{Supervised Fine-Tuning}
\acrodef{CoT}[CoT]{Chain-of-Thoughts}
\acrodef{VS}[VS]{Vendi Score}
\acrodef{MoE}[MoE]{Mixture of Experts}
\acrodef{Q-D}[Q-D]{Quality-Diversity}
\title{Synthetic Data Generation for Training\\ Diversified Commonsense Reasoning Models}

\author{
Tianhui Zhang$^{*}$ \quad
Bei Peng$^{\dagger}$ \quad
Danushka Bollegala$^{*,\diamondsuit}$ \\
$^{*}$University of Liverpool \quad
$^{\dagger}$University of Sheffield \quad $^{\diamondsuit}$Amazon\\
\texttt{\{tianhui.zhang, danushka\}@liverpool.ac.uk \quad bei.peng@sheffield.ac.uk}
}

\date{}

\begin{document}
\maketitle

\begin{abstract}
    Conversational agents are required to respond to their users not only with high quality (i.e. commonsense-bearing) responses, but also considering multiple plausible alternative scenarios, reflecting the diversity in their responses.
    Despite the growing need to train diverse commonsense generators, the progress of this line of work has been significantly hindered by the lack of large-scale high-quality diverse commonsense training datasets.
    Due to the high annotation costs, existing \ac{GCR} datasets are created using a small number of human annotators, covering only a narrow set of commonsense scenarios.
    To address this training resource gap, we propose a two-stage method to create \textbf{CommonSyn}, the first-ever synthetic dataset for diversified \ac{GCR}.
    \acp{LLM} fine-tuned on CommonSyn show simultaneous improvements in both generation diversity and quality compared with vanilla models and models fine-tuned on manually annotated datasets.\footnote{The dataset and code are available \href{https://github.com/LivNLP/CommonSyn}{here}.}
\end{abstract}

\section{Introduction}
\label{sec:intro}
\begin{figure}[t]
  \centering
  \includegraphics[width=0.95\columnwidth]{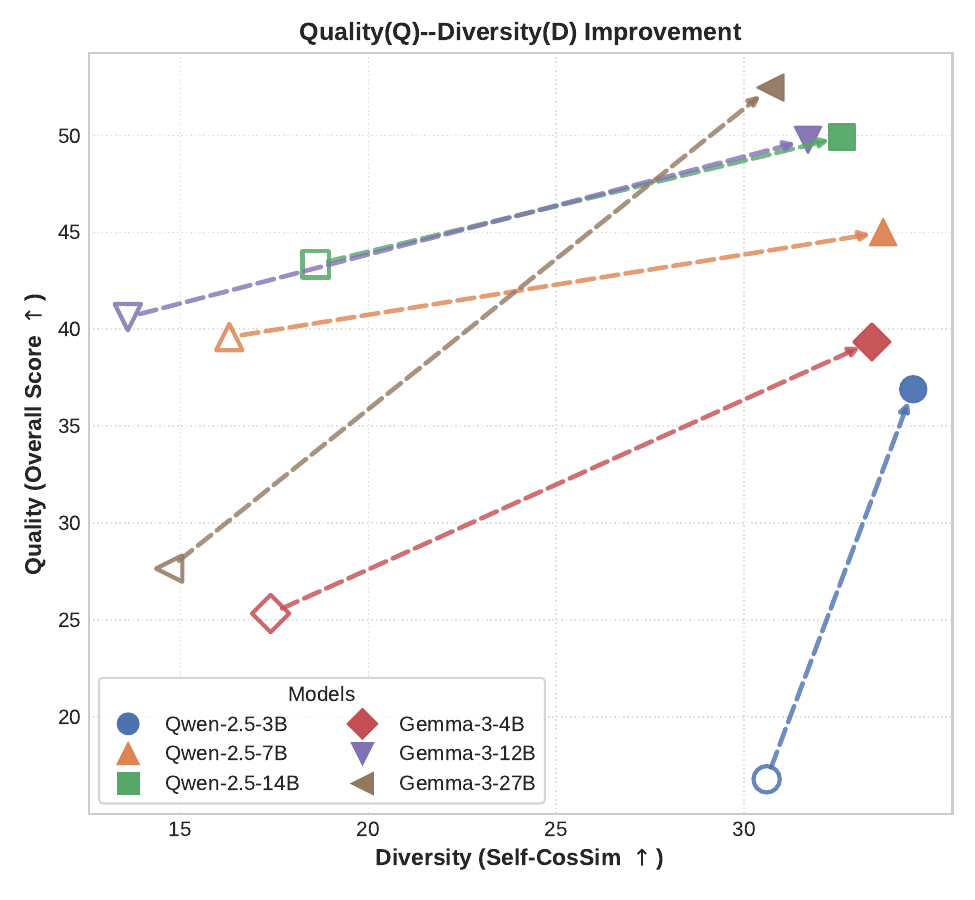}
  \caption{
  \textbf{Quality--Diversity trade-off for representative models.}
  The x-axis represents the semantic diversity (Self-CosSim, $\uparrow$) and the y-axis represents the generation quality (Overall, $\uparrow$).
  While vanilla models (hollow markers) often suffer from either low quality or limited diversity, its fine-tuned version on our synthetic data, \textsc{CommonSyn} (solid markers),
  consistently pushes the performance frontier towards the top-right quadrant, achieving a superior Pareto improvement across diverse model families.
  }
  \label{fig:quality_diversity}
\end{figure}

\ac{GCR} requires models to produce diverse, high-quality sentences from limited concepts~\cite{CommonGen,dimongen}.
In \ac{GCR}, a \ac{NLG} model is expected to generate sentences that are both \emph{quality-bearing} (i.e. logically coherent and commonsense-aware) and \emph{diverse} (i.e. offering varied scenarios) based on limited input information (i.e. few concepts) relying on the models' intrinsic or externally retrieved knowledge~\citep{dimongen,yu:2022:diversifying,hwang:2023:knowledge}.
Although CommonGen~\cite{CommonGen} serves as a standard benchmark for \ac{GCR}, it suffers from critical data limitations: (1) \textbf{limited diversity}, as 83\% of concept sets have fewer than two references; and (2) \textbf{quality issues}, including near-duplicates and incomplete structures. 
Consequently, models fine-tuned on existing datasets often struggle to balance quality with semantic diversity or generalise to unseen concepts.

Synthetic data generation offers a scalable solution~\citep{yu2024metamath, guo2024deepseek}. 
Compared to human annotations, it is easier to obtain scalable and controllable quality data using \acp{LLM}~\citep{bauer2024comprehensive, liu2024best}. 
However, recent studies have shown that synthetic data often face challenges such as low structural diversity, strong stylistic biases, and model collapse~\citep{yu2023large,shumailov2023curse}.
Although recent work has attempted to diversify synthetic data~\citep{yu2023large,miranda2023beyond,ge2024scaling}, they mainly optimise the diversity of the \emph{data itself} rather than the \ac{Q-D} trade-off of the fine-tuned model-output. 
It remains unclear how to create synthetic data for \ac{GCR} so that both quality and diversity improve simultaneously after fine-tuning.

% Although data augmentation techniques have been proposed for \ac{GCR}~\citep{zhang:2024:diversity}, existing work primarily focusses on improving the diversity \emph{of the synthetic data itself}, but does not answer the key question: \emph{what properties of a synthetic dataset guarantee that a model fine-tuned on that dataset improves both quality and diversity simultaneously for \ac{GCR}}.
To bridge the gap, we propose \textbf{CommonSyn}, the first-ever synthetic dataset specifically designed for diversifying \ac{GCR} considering CommonGen~\citep{CommonGen}, a widely-studied \ac{GCR} setup.
% Our goal is not only to scale the CommonGen-style training data, but more importantly to identify which data generation and selection strategies push the Quality–Diversity trade-off under fine-tuning.
We propose a two-stage method that evaluates multiple concept expansion and sample selection strategies to identify the optimal \ac{Q-D} frontier. 
By balancing the local diversity within each concept set and the global diversity across the entire dataset, CommonSyn not only improves the quality but also the diversity of several \acp{LLM} when fine-tuned, compared to using the manually annotated CommonGen training set. 
Furthermore, those fine-tuned models further improve performance of several other \ac{GCR} tasks, demonstrating the generalisability of CommonSyn.

Our main contributions are as follows:
\begin{itemize}
    \item We propose \textbf{CommonSyn}, a synthetic dataset that enables fine-tuned models to consistently outperform those trained on human-annotated data in both quality and diversity.
    \item We propose a 2-stage data selection method that balances local diversity within each concept set as well as global diversity throughout the dataset, which is computationally efficient compared to gradient-based selection methods~\cite{jung2025prismatic} that require computing loss gradients over model parameters.
    \item Extensive experiments across 11 \acp{LLM} demonstrate robust Pareto improvements. Furthermore, models fine-tuned on CommonSyn generalise well to related tasks such as abductive reasoning and story completion.
\end{itemize}

\section{Related Work}
\label{sec:related}
\paragraph{Diversifying the GCR.} % I think we should focus on CommonGen because this is where we create the synthetic data. Existing methods only improve diversity of model outputs, not dataset quality-diversity
To diversify \ac{GCR}, a model must generate both commonsense-bearing and diverse sentences. Datasets such as CommonGen~\citep{CommonGen} provide a set of concepts and a set of sentences that describe various commonsense relations among those concepts, while ComVE~\citep{semeval} requires a \ac{GCR} method to explain why a given counterfactual statement does not make commonsense. % Here we also can put DimonGen and α-NLG ( Abductive commonsense reasoning dataset t (Bhagavatula et al., 2020))
Methods for improving diversity in \ac{GCR} have been explored from various angles. 
Sampling-based decoding strategies, such as nucleus sampling~\citep{holtzman:2019:sample} improve diversity by sampling next tokens from the nucleus of the generation probability distribution.
Other approaches integrate external information from a knowledge graph~\citep{hwang:2023:knowledge} or retrieve sentences from large corpora~\citep{dimongen,cui2024more} to increase the diversity of the generation.
\citet{zhang:2024:diversity} used human-written prompts to explicitly instruct \acp{LLM} to generate diverse outputs.
However, to the best of our knowledge, existing methods focus only on improving the diversity of model outputs, neither creating synthetic training data nor addressing the challenge of jointly improving both generation quality and diversity.

\paragraph{Synthetic Data Creation.}
% -------------------------------------------
Synthetic data refers to artificially generated data that mimics the characteristics of real-world data~\citep{liu2024best}.
\acp{LLM} have been used for the generation of synthetic data to augment questions or solutions~\citep{jung2025prismatic, yu2024metamath}.
Previous studies used synthetic data for fine-tuning~\citep{huang2025key,chen2024self}, instruction tuning~\citep{li2024culturellm, wu2024lamini}, knowledge transfer~\citep{meng2022generating} and evaluation~\citep{zhumeasuring}.
To avoid the model collapse problem (i.e., generating highly similar data leading to diminishing model performance), methods have been proposed to diversify synthetic data~\citep{feng2025beyond}. 
\citet{ge2024scaling} generated various prompts with customised fine-grained personas.
\citet{wang-etal-2025-diversity} augmented the training data by training a diversification model.
\citet{jung2025prismatic} generated data based on the gradient vector of a proxy model. 
Although methods for diversifying synthetic data have been proposed~\citep{chen2024diversity, zhumeasuring}, they do not evaluate the joint \ac{Q-D} of the fine-tuned model's outputs.

\paragraph{Data Selection.}
Data selection is the task of selecting a subset of desirable training samples from a larger training dataset, where automatic data selection methods have been proposed~\citep{sener2017active, kaushal2019learning, xia2022moderate}.
According to~\citet{qin2024unleashing}, existing strategies can be categorised into three groups: (a) \textbf{Quality-based} methods that filter data using statistical indicators such as perplexity~\citep{ankner2024perplexed} or \ac{LLM}-based scoring~\citep{chen2023alpagasus} to ensure correctness, 
(b) \textbf{Diversity-based} methods that aim to reduce redundancy or improve coverage by clustering or sampling from the embedding space~\citep{sener2017active, tirumala2023d4, xie2023data}, and
(c) \textbf{Importance-based} methods identify influential samples that significantly impact model parameters using gradient information~\citep{deng2024influential,pan2024g,jung2025prismatic}.
\textbf{However, to the best of our knowledge, we are the first to create a synthetic dataset by combining those selection strategies to jointly improve quality and diversity in \ac{GCR}.}

\section{Synthetic Data Generation for GCR}
\label{sec:method}
\subsection{Problem Setting}
\ac{GCR} tasks such as CommonGen~\citep{CommonGen} require a model to produce a natural, coherent, and plausible description of a scenario given a set of 3-5 concepts. 
For example, given the concept set ``\emph{bicycle}, \emph{ride}, \emph{street}'', a plausible sentence could be ``\emph{The woman rides a bicycle down the street instead of walking.}''
Formally, given a concept set $\cX = \{x_1, \ldots, x_m\}$, the goal is to generate a sentence $y$ that covers all concepts in $\cX$ while maintaining grammatical correctness and commonsense plausibility.
% What is diversified GCR
Beyond producing a single sentence, the diversified \ac{GCR} aims to generate a set of sentences $\cY= \{y_1,  \ldots, y_n\}$, such that each individual sentence is \emph{quality-bearing}, while
the set of generated sentences collectively exhibits  \emph{diversity}~\cite{li:2016:distinct}.
Unlike decoding strategies that diversify output during inference~\cite{holtzman:2019:sample}, we focus on the \textbf{training data}.
We construct a synthetic dataset named \textbf{CommonSyn}, $\cD_{syn} = \{(\cX_i, \cY_i)\}_{i=1}^N$, designed to improve the Pareto frontier of quality and diversity for models fine-tuned on it, resulting in a better quality-diversity trade-off on unseen test sets.

\subsection{Concept Set Generation}
\label{sec:method_concept}
\begin{table}[t]
\centering
\resizebox{\columnwidth}{!}{%
\begin{tabular}{lcc}
\toprule
\textbf{Statistics} & \textbf{CommonSyn} & \textbf{CommonGen Train} \\
\midrule
\# Concept-Sets & 20, 742 & 32,651 \\
\quad -- Size = 3 & 9,847 & 25,020 \\
\quad -- Size = 4 & 6,857 & 4,240 \\
\quad -- Size = 5 & 4,038 & 3,391 \\
\addlinespace
\# Sentences & 83,184 & 67,389 \\
per Concept-Set & 4.01 & 2.06 \\
Average Length & 14.19 & 10.54 \\
\bottomrule
\end{tabular}%
}
\caption{Comparison between our synthetic dataset CommonSyn and the CommonGen training set in terms of scale and structure.}
\label{tab:data_comparison}
\end{table}
Constructing high-quality concept sets for synthetic GCR is challenging as it must satisfy two conflicting criteria:
(i) \textbf{Coherence:} concepts must be sufficiently semantically related to form a plausible daily scenario, and
(ii) \textbf{Coverage:} the concept sets should cover a wide range of unique concepts and combinations to prevent a model from over-fitting to a restricted body of knowledge.
The original CommonGen concept sets are limited in scale and sentences per concept set (\autoref{tab:data_comparison}), but they provide an important structural insight: concepts appearing together typically lie within 2-hop distance in ConceptNet~\citep{speer2017conceptnet}, indicating strong semantic and commonsense relations.
Therefore, we use this property to design a simple yet effective concept expansion procedure.

Instead of using complete concept sets only from the human-annotated CommonGen training data (which limits diversity) or generating concepts from scratch (which often yields unrelated terms), we use the existing CommonGen training data as a candidate set of seeds.
Specifically, given a concept set from the CommonGen training set, we randomly sample two concepts as \emph{anchors} to instruct an \ac{LLM} to generate 1-3 additional bridge concepts to form a novel, commonsense-bearing scenario (prompt details in \autoref{fig:synthetic_prompt}).
This method ensures that the generated concepts maintain the semantic compatibility encoded in CommonGen while exploring novel combinations of concepts.
For example, given the original concept set $\{$\emph{bicycle}, \emph{ride}, \emph{street}$\}$, we randomly select \emph{bicycle} and \emph{ride} as anchors. 
In this case, \ac{LLM} might generate \emph{path} as an additional concept, resulting in a new concept set. 
This approach is based on the observation that in the original CommonGen~\citep{CommonGen} 50\% of concept sets are fully connected within a distance of 2-hops, ensuring semantic coherence in the expanded concept set.

\subsection{Sentence Generation}
\label{sec:sentence-generation}

% TZ: This section we will introduce the three sources (dynamic/fewshot/CoT), and we will also say the dynamic does not work. Then we can introduce their advantages and disadvantage of each method: FS → high diversity Dynamic → high quality CoT → descriptive reasoning
% And why we donlt use CoT to diversifying output.
%We find that naive scaling of synthetic data does not improve the model, facing early saturation even with heuristic diversification such as persona-guided prompting~\citep{ge2024scaling}
% -------------------------------------------
Given the expanded concept sets, we use three distinct prompting strategies to generate candidate sentences. 
Each strategy targets a different region in the \ac{Q-D} space (see \autoref{tab:source_ablation}), producing $N$ candidates per concept set\footnote{We use 
$N=4$, which is in agreement with the number of reference sentences included in CommonGen test sets.}.
All outputs are limited to a maximum of 22 words to match the task distribution.
The prompts used for this generation step are shown in~\autoref{app:sec:gen_prompt}.

\begin{itemize}[leftmargin=*]
    \item \textbf{Dynamic Few-Shot Generation ($D_{dyn}$)}: We generate sentences one at a time.
For each generation, we randomly sample $k=5$ few-shot examples from the CommonGen training set. 
Although this follows the standard prompting protocol~\citep{CommonGen}, we observe that despite the randomised examples, the \ac{LLM} tends to generate highly similar sentences. 
    
    \item \textbf{Multi-sentence Few-shot Generation ($D_{ms}$)}: This strategy also uses few-shot prompting, but unlike $D_{dyn}$, we instruct the \ac{LLM} to produce $N$ \emph{different} candidate sentences within a single generation step~\citep{zhang:2024:diversity}.
    By placing multiple outputs in the same context window, the model is implicitly encouraged to diversify subsequent sentences to avoid repetition. 
    This strategy improves both lexical and syntactic variation, although the quality may fluctuate compared to the single step generation approach.
    
    \item \textbf{\ac{CoT}-guided Generation ($D_{cot}$)}: \ac{CoT} has significantly improved the reasoning abilities of \acp{LLM} in a wide range of tasks~\citep{wei2022chain, naik2023diversity}. 
    To capture deeper semantic relations between concepts, we prompt the \ac{LLM} to first produce a short explanation describing how the input concepts might relate to each other, and then require it to output the desired sentences one at a time.
    This reasoning step encourages the model to explicitly consider causal or temporal relations before composing a sentence, leading to different styles of candidate sentences compared to those generated with $D_{dyn}$ or $D_{ms}$.
\end{itemize} 

After the generation step, we apply a \textbf{keyword coverage filter} to ensure that all input concepts appear in the output sentences.
We combine all remaining sentences into a unified candidate pool.
As shown later in \autoref{sec:exp:source_gen}, naïvely merging the candidates (12 sentences per concept set) from the three strategies does not improve the Q–D characteristics of the fine-tuned model, calling for a principled data selection approach.

\begin{table*}[t]
\centering

\resizebox{0.9\textwidth}{!}{
\begin{tabular}{ll ccc c cccc}
\toprule
\multirow{2}{*}{\textbf{Model}} & \multirow{2}{*}{\textbf{Dataset}} & \multicolumn{3}{c}{\textbf{Quality Metrics} ($\uparrow$)} & & \multicolumn{4}{c}{\textbf{Diversity Metrics} ($\uparrow$)} \\
\cmidrule{3-5} \cmidrule{7-10}
 & & Win-Tie & Cov. & \textbf{Overall} & & S-BLEU3$^\dagger$ & S-BLEU4$^\dagger$ & Vendi & \textbf{S-Cos}$^\dagger$ \\
\midrule
\multirow{3}{*}{Llama-3.1-8B-Inst} 
 & Vanilla & 19.0 & 84.7 & 16.1 & & 73.4 & 80.3 & \textbf{22.8} & \textbf{35.1} \\
 & CommonGen & 31.7 & 85.8 & 27.2 & & \textbf{76.4} & \textbf{84.7} & 21.2 & 30.7 \\
 & \textbf{CommonSyn} & \textbf{47.3\textsuperscript{**}} & \textbf{94.5} & \textbf{44.7} & & 75.5 & 82.9 & 22.7 & 34.2 \\
\midrule
\multirow{3}{*}{Qwen-2.5-7B-Inst.} 
 & Vanilla & 42.9 & 92.3 & 39.6 & & 39.2 & 44.1 & 15.8 & 16.3 \\
 & CommonGen & 32.5 & 88.6 & 28.8 & & 73.2 & \textbf{81.8} & 21.4 & 31.2 \\
 & \textbf{CommonSyn} & \textbf{48.4\textsuperscript{**}} & \textbf{93.0} & \textbf{45.0} & & \textbf{73.7} & 81.0 & \textbf{22.5} & \textbf{33.7} \\
\midrule
\multirow{3}{*}{Qwen-2.5-14B-Inst.} 
 & Vanilla & 45.2 & 95.9 & 43.4 & & 42.4 & 47.5 & 16.5 & 18.6 \\
 & CommonGen & 38.5 & 92.9 & 35.8 & & 70.8 & 79.6 & 20.9 & 29.7 \\
 & \textbf{CommonSyn} & \textbf{52.0\textsuperscript{**}} & \textbf{96.0} & \textbf{49.9} & & \textbf{73.2} & \textbf{80.6} & \textbf{22.1} & \textbf{32.6} \\
\midrule
\multirow{3}{*}{Gemma-3-4B-IT} 
 & Vanilla & 28.7 & 88.4 & 25.3 & & 37.7 & 42.3 & 16.1 & 17.4 \\
 & CommonGen & 28.7 & 91.0 & 26.1 & & 74.9 & \textbf{83.2} & 21.2 & 30.6 \\
 & \textbf{CommonSyn} & \textbf{41.2\textsuperscript{**}} & \textbf{95.5} & \textbf{39.4} & & \textbf{75.4} & 82.6 & \textbf{22.4} & \textbf{33.4} \\
\midrule
\multirow{3}{*}{Gemma-3-12B-IT} 
 & Vanilla & 43.2 & 94.0 & 40.6 & & 31.7 & 35.7 & 14.7 & 13.6 \\
 & CommonGen & 35.9 & 94.1 & 33.8 & & 72.2 & 80.4 & 20.5 & 28.7 \\
 & \textbf{CommonSyn} & \textbf{51.1\textsuperscript{**}} & \textbf{97.4} & \textbf{49.8} & & \textbf{73.7} & \textbf{81.2} & \textbf{21.8} & \textbf{31.7} \\
\midrule
\multirow{3}{*}{Gemma-3-27B-IT} 
 & Vanilla & 31.9 & 86.7 & 27.7 & & 27.0 & 30.2 & 15.0 & 14.7 \\
 & CommonGen & 40.3 & 95.6 & 38.5 & & 70.7 & 79.2 & 20.3 & 28.1 \\
 & \textbf{CommonSyn} & \textbf{53.6\textsuperscript{**}} & \textbf{97.9} & \textbf{52.5} & & \textbf{71.7} & \textbf{79.2} & \textbf{21.4} & \textbf{30.7} \\
\bottomrule
\end{tabular}
}
\caption{Main results for the top-6 performing models. When fine-tuned on our proposed dataset (\textbf{CommonSyn}), we observe consistent improvements achieving the best balance between quality and diversity, surpassing both the vanilla model (prior to fine-tuning) and human-annotated CommonGen data (\textsuperscript{**}: $p<0.01$). The complete results are shown in~\autoref{tab:full_results}. Diversity metrics marked with $\dagger$ are inverted by subtracting from 1.}
\label{tab:main_results}
\end{table*}

\subsection{Data Selection}
To construct the final \textbf{CommonSyn} dataset, we use a two-stage method that balances \emph{local} diversity (within a concept set) and \emph{global} quality/coverage.
% To address this issue, we design a two-stage Q–D selection method, consisting of a \emph{local} filtering stage and a \emph{global} re-ranking stage.
% This design considers the requirements of diversified CommonGen: diversity must be preserved within each concept set (local), while the final dataset must remain globally balanced in terms of quality and semantic coverage.

% \begin{figure}[h]
% \centering
% \footnotesize
% \begin{minipage}{\columnwidth}
% \setlength{\parskip}{2pt}

% \textbf{Quality Scoring Prompt:} \\
% \fbox{\parbox{\dimexpr\linewidth-2\fboxsep-2\fboxrule\relax}{
% You are an expert evaluator of sentence quality and commonsense reasoning.

% Score each sentence 1–10 (integer values only). Guidelines: \\
% 1–3: Implausible, ungrammatical, or missing concepts. \\
% 4–6: Minor clarity or concepts-use issues. \\
% 7–8: Clear, fluent, plausible, good integration. \\
% 9–10: Exceptional realism and flawless integration. \\
% Assign score 1 to ``[EMPTY]'' sentences.

% Evaluate for commonsense plausibility, meaningful concept coverage, grammatical clarity.

% Concept set: \{concept\_set\} \\
% Candidates: \{sentence\_list\}
% }}

% \end{minipage}
% \caption{The prompt used for score per-sentence plausibility ($Q(y) \in [1,10]$) and filter out low-quality candidates ($Q<4$). Further details given in~\autoref{fig:prompt:quality_scorer_detailed}.}
% \label{fig:prompt:quality_scorer}
% \end{figure}

\paragraph{Local Selection.}
\label{para:local_select}
For each concept set $\cC$, we first filter out low-quality candidates.
We use a lightweight scorer (Gemini-2.5-flash) to assign a plausibility score $Q(y) \in [1,10]$ (with 10 indicating the most plausible scenarios) to each sentence, scoring for its grammatical correctness and commonsense validity (\autoref{fig:prompt:quality_scorer_detailed}).
Then we discard those with $Q(y)<4$.
% For each concept set $\cC$, we first evaluate the quality of each generated candidate sentence.
% We employ a \emph{quality scorer} (i.e., Gemini-2.5-flash) and assign a plausibility score $Q(y)$ in the range of 1-10 (with 10 indicating the most plausible scenarios) to each sentence $y$, scoring for its grammatical correctness and commonsense validity (\autoref{fig:prompt:quality_scorer_detailed}).
% To remove low-quality candidates, we select sentences with a minimal $Q(y)$ of 4.

For the remaining candidates $\mathcal{Y}_{\mathcal{C}}$ within the same concept set, we compute their average pairwise cosine similarity using SimCSE sentence embeddings~\citep{gao:2021:simcse}.
Specifically, for a sentence $y_i$, its local diversity score, $D_{\text{local}}(y_i)$, is defined as:
\begin{align}
    D_{\text{local}}(y_i) = 1 - \frac{1}{{|\mathcal{Y}_{\mathcal{C}}|-1}} \sum_{y_j \in \mathcal{Y}_{\mathcal{C}}, j \ne i} \cos(\mathbf{e}_i, \mathbf{e}_j) .
\end{align}
We then select the top–$k$ sentences with the highest $D_{\text{local}}$ scores for each concept set (setting $k=8$ in our experiments) to maximise the \textbf{local diversity}. 
This step ensures that our synthetic dataset maintains high intra-concept-set variety.

\paragraph{Global Selection.}
We aggregate all locally selected candidates into a global pool $\cP$.
To ensure the final dataset is not only locally diverse but also globally balanced, we compute a global diversity score $D_{\text{global}}(s)$ for each sample $s$ against the entire pool.
Let $g = \sum_{s' \in \cP} e_{s'}$ denote the sum of all embedding vectors, where each embedding vector is $\ell_2$-normalised, i.e. $\norm{\vec{e}_s}_2 = 1$.
We define the global diversity score as:
\begin{align}
    D_{\text{global}}(s) = 1 - \frac{{\vec{e}_s}\T(\vec{g} - \vec{e}_s)}{|\cP| - 1}
\end{align}
Finally, we rank all samples using a joint score. To ensure comparability, both $Q(s)$ and $D_{\text{global}}(s)$ are min-max normalised:
\begin{align}
    S(s) = \tilde{Q}(s) + \tilde{D}_{\text{global}}(s),
\end{align}
and select the top scoring samples (i.e. 83,184) as our final synthetic dataset, CommonSyn.
This global ranking process balances quality and semantic coverage across the entire dataset, encouraging the model to learn not only plausible, but also diverse sentence patterns.

We also experimented with gradient-based data selection where we cluster training candidates in the gradient space of a proxy model and up-sample from the sparsest clusters 
to maximise the diversity of training influences~\citep{pan2024g, jung2025prismatic}.
However, empirical results (\autoref{tab:selection_ablation}) indicated that even combining gradient with $Q(s)$ does not yield a better \ac{Q-D} trade-off than the proposed embedding-based method.
Furthermore, gradients are model-dependent and computationally expensive to compute (computed over all model parameters). 

\begin{table}[ht]
\centering
\small
\setlength{\tabcolsep}{3pt} 
\resizebox{\columnwidth}{!}{
\begin{tabular}{l@{\hspace{1mm}} ccc ccc}
\toprule
\multirow{2}{*}{\textbf{Method}} & \multicolumn{3}{c}{\textbf{Quality} ($\uparrow$)} & \multicolumn{3}{c}{\textbf{Diversity} ($\uparrow$)} \\
\cmidrule(lr){2-4} \cmidrule(l){5-7}
 & Win-Tie & Cov. & \textbf{Overall} & S-B4$^\dagger$ & Vendi & \textbf{S-Cos}$^\dagger$ \\
\midrule
\multicolumn{7}{l}{\textit{\textbf{Baselines}}} \\
CommonGen & 31.7 & 85.8 & 27.2 & 84.7 & 21.2 & 30.7 \\
PersonaHub~\citep{ge2024scaling} & 20.8 & 94.0 & 19.6  & 78.1 & \textbf{23.8} & 35.6 \\
Prismatic~\citep{jung2025prismatic} & 44.2 & 91.6 & 40.4  & 82.9 & 21.5 & 31.1 \\
\addlinespace
\midrule
\multicolumn{7}{l}{\textit{\textbf{Sentence Generation Methods}}} \\
Few-shot ($D_{ms}$) & 44.6 & 90.5 & 40.4 & \textbf{85.5} & 23.4 & \textbf{36.3} \\
Dynamic ($D_{dyn}$) & \textbf{53.1} & 90.9 & \textbf{48.3} & 68.5 & 19.1 & 24.6 \\
CoT ($D_{CoT}$) & 43.4 & 95.0 & 41.2 & 80.6 & 21.9 & 32.0 \\
\addlinespace
\midrule
\multicolumn{7}{l}{\textit{\textbf{Pooled Candidates}}} \\
2 best ($D_{dyn} \cup D_{ms}$) & 50.0 & \textbf{95.3} & 47.7 & 75.7 & 21.0 & 30.0 \\
All ($D_{dyn} \cup D_{ms} \cup D_{CoT}$) & 50.0 & 94.4 & 47.2 & 78.8 & 21.6 & 31.4 \\
\addlinespace
\midrule
\multicolumn{7}{l}{\textit{\textbf{Selection Strategy}}} \\
\textbf{CommonSyn} & 47.3 & 94.5 & 44.7 & 82.9 & 22.7 & 34.2 \\
\bottomrule
\end{tabular}
}
\caption{Comparison of different data sources and selection strategies on Llama-3.1-8B. \textbf{CommonSyn} effectively balances quality and diversity, maintaining high Coverage. Metrics marked with $\dagger$ are inverted (by subtracting from 1) such that higher values indicate better performance.}
\label{tab:source_ablation}
\end{table}

\section{Experiments}
\label{sec:exp}

\subsection{Implementation Details}
\label{sec:implement}
For our synthetic data, we use \href{https://huggingface.co/Qwen/Qwen2.5-72B-Instruct}{Qwen2.5-72B-Instruct} with $T=1.0$ to generate both concept sets and the corresponding sentence sets with the three sentence generation strategies for reproducibility.
The 2-Seed Expansion strategy uses two randomly selected concepts per seed to generate novel combinations. 
To prevent potential data leakage, we ensure that concepts from CommonGen test sets are excluded from the training set, resulting in a 99.6\% unseen concept-triple rate (see~\autoref{tab:concept_construction}).
For Dynamic Few-shot, we use 5 exemplars with temperature set to $T=1.0$. 

We selected a total of 11 widely used instruct-tuned models for fine-tuning and evaluation: 
\href{https://huggingface.co/meta-llama/Llama-3.1-8B-Instruct}{Llama-3.1-8B}, 
\href{https://huggingface.co/meta-llama/Llama-3.2-1B-Instruct}{3.2-1B}, 
\href{https://huggingface.co/meta-llama/Llama-3.2-3B-Instruct}{3B};
\href{https://huggingface.co/Qwen/Qwen2.5-1.5B-Instruct}{Qwen-2.5-1.5B},
\href{https://huggingface.co/Qwen/Qwen2.5-3B-Instruct}{3B},
\href{https://huggingface.co/Qwen/Qwen2.5-7B-Instruct}{7B},
\href{https://huggingface.co/Qwen/Qwen2.5-14B-Instruct}{14B};
\href{https://huggingface.co/google/gemma-3-1b-it}{Gemma3-1B},
\href{https://huggingface.co/google/gemma-3-4b-it}{4B},
\href{https://huggingface.co/google/gemma-3-12b-it}{12B},
\href{https://huggingface.co/google/gemma-3-27b-it}{27B}.
We fine-tuned these models under \ac{SFT}.  
All models are fine-tuned with Unsloth~\citep{unsloth} using LoRA with rank $r=96$ and $\alpha=192$. 
The prompt for training samples is shown in~\autoref{tab:commonsyn_prompts}.
For inference, we use a temperature of $T=1.0$ to balance diversity and coherence (higher $T$ encourages more diverse outputs).
For data selection, we use \textbf{Gemini-2.5-Flash} as the quality scorer because of its cost-efficiency.

To compare our method with gradient-based data selection on the Llama-3.1-8B model, we follow the procedure of \citet{jung2025prismatic}. 
Specifically, we compute gradients of candidate data samples on the model with respect to all trainable parameters, flatten and concatenate them into a single vector. 
Then we randomly project the resulting high-dimensional gradients into a 1024-dimensional space using TRAK \citep{park2023trak}.
We use K-means clustering algorithm and randomly sample sentences from the top-50\% rarest clusters to ensure the diversity of training data influences.
In practice, we cluster the training data into 831 clusters.

For the final evaluation, following the evaluation setting of~\cite{CommonGen}, we use GPT-4o to evaluate the preference between generated sentences and human-written references.
In both cases, the temperature is set to $T=0$ to obtain reproducible generations.
We used A100 GPUs for all experiments.

\begin{table}[t]
\centering
\resizebox{\columnwidth}{!}{
\begin{tabular}{p{3cm} p{10cm}}
\toprule
\textbf{Role} & \textbf{Prompt} \\
\midrule
System & Given several keywords, generate one coherent sentence that contains all the required keywords using background commonsense knowledge:  \\
\midrule
User & \{Concept Set\}  \\
\midrule
Assistant & \{Sentence\}\\
\bottomrule
\end{tabular}}
\caption{Prompts used for training samples for CommonSyn.}
\label{tab:commonsyn_prompts}
\end{table}

\subsection{Evaluation Protocol}
\label{sec:metrics}

% Fine tuning details
To evaluate the usefulness of CommonSyn, we fine-tune multiple \acp{LLM} on it and measure the improvements of \ac{Q-D} of the generated sentences.
Specifically, we conduct parameter-efficient \ac{SFT} using unsloth~\citep{unsloth} with LoRA for all models. 
We train the model to generate one sentence per training instance given its concept set and use the next token prediction as the training objective for \ac{SFT}.
We use the zero-shot prompt shown in~\autoref{tab:commonsyn_prompts} for this purpose.
Multiple quality and diversity metrics are applied in our evaluations.

\paragraph{Quality Metrics.}
Following the standard CommonGen protocol~\citep{CommonGen}, we assess quality using three metrics.
(i) \textbf{Coverage:} The percentage of test examples where the generated sentences include all of the input concepts.
(ii) \textbf{Win/Tie Rate:} We use GPT-4o as a judge to compare model outputs against human-written references, where the model outputs and references are randomly shuffled to ensure fairness. 
The prompt instructs the \ac{LLM} judge to select the sentence that is more natural, coherent, and satisfies the length constraint. 
To evaluate the reliability of the LLM-as-a-judge, we further conduct a human evaluation, which shows 81.6\% agreement with GPT-4o (see~\autoref{app:sec:eval_prompt} for details).

(iii) \textbf{Overall:} Defined as the product of Coverage and Win/Tie Rate, serving as our primary quality metric.

\paragraph{Diversity Metrics.}
To measure the lexical and semantic variation of the generated outputs, we use the following three metrics.
(i) \textbf{Self-BLEU}~\citep{zhu:2018:selfbleu} measures the $n$-gram overlap between the generated sentences. 
We used self-BLEU-3/4 (i.e. $n=3, 4$) in our experiments. 
(ii) \textbf{Self-CosSim}~\citep{cox2021directed} is the average pairwise cosine similarity of sentence embeddings.
We subtract the self-BLEU and self-CosSim scores by $1$, such that higher scores indicate greater pairwise diversity.
(iii) \textbf{\ac{VS}-embed}~\citep{vendi:2022} is the exponential of Shannon's entropy over the eigenvalues of the pairwise similarity matrix of a set of sentences computed using sentence embeddings.
We use the SimCSE~\citep{gao:2021:simcse} for both Self-CosSim and VS-embed for consistency.
Following ~\citet{zhang2025evaluating} we use Self-CosSim as our primary diversity metric because of its reliability.

\begin{table}[t]
\centering
\small
\setlength{\tabcolsep}{3.5pt}
\resizebox{\columnwidth}{!}{
\begin{tabular}{l l c ccc}
\toprule
\multirow{2}{*}{\textbf{Task}} &
\multirow{2}{*}{\textbf{Model}} &
\multicolumn{1}{c}{\textbf{Quality} ($\uparrow$)} &
\multicolumn{3}{c}{\textbf{Diversity} ($\uparrow$)} \\
\cmidrule(lr){3-3} \cmidrule(lr){4-6}
 & & \textbf{Win-Tie} & S-B4$^\dagger$ & Vendi & \textbf{S-Cos$^\dagger$} \\
\midrule
\multirow{3}{*}{ComVE} 
 & Vanilla & 75.3 & 77.3 & 17.7 & 29.8 \\
 & CommonGen & 41.7 & \textbf{86.3} & \textbf{21.6} & \textbf{44.9} \\
 & \textbf{CommonSyn} & \textbf{78.9} & 85.9 & 18.1 & 30.0 \\
\midrule
\multirow{3}{*}{$\alpha$-NLG} 
 & Vanilla & 54.6 & 83.5 & 18.8 & 24.2 \\
 & CommonGen & 25.9 & 81.3 & \textbf{22.6} & \textbf{35.3} \\
 & \textbf{CommonSyn} & \textbf{61.1} & \textbf{87.9} & 19.6 & 25.7 \\
\midrule
\multirow{3}{*}{ROCStories} 
 & Vanilla & 87.2 & 79.3 & 23.9 & 38.8 \\
 & CommonGen & 43.0 & \textbf{95.0} & \textbf{29.2} & \textbf{54.1} \\
 & \textbf{CommonSyn} & \textbf{88.6} & 90.8 & 24.1 & 38.6 \\
\bottomrule
\end{tabular}
}
\caption{Cross-task generalization on unseen generative commonsense tasks. Models fine-tuned on human data (\textit{CommonGen}) suffer from severe \textbf{quality drop}. In contrast, model fine-tuned on \textbf{CommonSyn} generalises well, outperforming vanilla baselines while maintaining diversity. Metrics marked with $\dagger$ are inverted.}
\label{tab:gen_transfer}
\end{table}
\subsection{Main Results}
\label{sec:exp:main}
% We show the performance of the synthetic data on several different models showing that it is not a model-dependent synthetic data. 
% -------------------------------------------
We fine-tune several \acp{LLM} on CommonSyn and compare them against vanilla (zero-shot) and CommonGen-finetuned baselines as shown in \autoref{tab:main_results} (see \autoref{tab:full_results} for additional results).
Models trained on CommonSyn achieve significantly higher \textbf{Win-Tie} rates and \textbf{Coverage} than the vanilla and CommonGen baselines (significant at $p<0.01$ according to the bootstrap sampling test~\citep{hesterberg2011bootstrap}).
For example, although Qwen-2.5-7B/14B already has strong performance on quality metrics, CommonSyn boosts their \textbf{Overall} score (+5.4) and doubling the diversity scores (S-Cos 16.3 to 33.7), showing that CommonSyn successfully injects semantic diversity without degrading commonsense quality.

A notable exception in diversity is observed with Llama-3.1-8B-Inst, where the vanilla model exhibits high diversity scores (S-Cos 35.1) but extremely low quality on both \textbf{Coverage} and \textbf{Win-Tie}. 
As further analysed later in~\autoref{sec:exp:quantitive}, this stems from verbose generations that fail to reliably cover all of the given input concepts. 
In contrast, CommonSyn improves the \textbf{Overall} score (from 16.1 to 44.7), while not harming the diversity too much. This confirms that CommonSyn jointly optimises the \ac{Q-D} objective.
Moreover, CommonSyn shows consistent gains across model sizes ranging from 4B to 27B parameters.
The 27B model fine-tuned on CommonSyn achieves the highest reported \textbf{Win-Tie} rate, demonstrating that even a capable large model can be further improved for quality using diversity-aware synthetic fine-tuning.

\subsection{Sentence Generation Strategies}
\label{sec:exp:source_gen}
% When you combine few-shot, dynamic, and CoT into a large pool: Dynamic provides the highest Q (Quantitative) few-shot provides the highest D (Digital) CoT provides intermediate style/reasoning-like text No selection can dominate all three types of sources simultaneously. there is no free lunch in Q–D when mixing multiple generative sources
% -------------------------------------------

In \autoref{tab:source_ablation}, we compare the different sentence generation methods, pooled candidates and the selection strategy described in \autoref{sec:sentence-generation} by fine-tuning Llama-3.1-8B-Instruct.
%This decomposition reveals why a multi-source, two-stage selection approach is necessary.
We observe that each individual generation strategy has different \ac{Q-D} performance. 
Although \textbf{Dynamic Few-Shot Generation ($D_{dyn}$)} achieves the highest \textbf{Win-Tie} rate, it shows severe mode collapse, resulting in the lowest diversity.
On the other hand, \textbf{Multi-sentence Few-shot Generation ($D_{ms}$)} produces the most diverse outputs but suffers from both lower quality and concept set Coverage. 
Although \textbf{\ac{CoT}-guided Generation ($D_{CoT}$)} has a moderate Overall Score, it reports the highest \textbf{Coverage} (95.0\%) among all sentence generation methods due to explicit reasoning.
% By making its reasoning steps explicit, it connects semantically distant ideas more clearly, which helps avoid missing important keywords.

Given the complementary strengths of the three sentence generation strategies, we found that simply taking the union of their generated sentence sets to be suboptimal, as shown in \autoref{tab:source_ablation} under \textbf{Pooled Candidates}.
Although pooling improves \textbf{Coverage}, pooled data underperforms $D_{ms}$ in diversity and $D_{dyn}$ in quality.
This confirms that naive merging introduces noise that reduces the benefits of high-potential candidates, emphasising the importance of sentence selection.

We also compared two baselines, PersonaHub~\citep{ge2024scaling} and Prismatic~\citep{jung2025prismatic}.
\href{https://huggingface.co/datasets/proj-persona/PersonaHub}{PersonaHub} is a synthetic dataset containing diverse personas, automatically curated from web data.
We generated both concept sets and corresponding sentences by sampling different personas using the same generation model as CommonSyn. 
CommonSyn outperforms PersonaHub in quality and achieve comparable diversity. 
We find that the model fine-tuned on persona-generated data learns to produce role-playing-style sentences that attempt to both cover the input concepts and reflect the persona's background, which harms plausibility. 
CommonSyn also outperforms Prismatic Synthesis, a gradient-based synthetic data generation method which generates and selects candidates by clustering samples in the gradient space of the \ac{LLM} and up-samples according to a sparsity metric defined over the sentence clusters. 
Furthermore, Prismatic Synthesis needs proxy models and the computation of the gradients over all of the samples in an iterative manner, which is computationally expensive in real-time applications.
In contrast, our proposal remains more computationally attractive.
% Although prismatic synthesis outperforms the original CommonGen, it still falls short of CommonSyn in both quality and diversity.

\begin{table}[t]
\centering
\small
\setlength{\tabcolsep}{4pt}
\resizebox{0.9\columnwidth}{!}{
\begin{tabular}{l ccc c}
\toprule
\textbf{Dataset} & \textbf{CSQA} & \textbf{CSQA2} & \textbf{PIQA} & \textbf{Avg.} \\
\midrule
Vanilla & 70.3 & 51.8 & 69.2 & \textbf{63.8} \\
CommonGen & 53.6 & \textbf{60.4} & 67.0 & 60.1 \\
\textbf{CommonSyn} & \textbf{70.8} & 49.0 & \textbf{71.3} & 63.7 \\
\bottomrule
\end{tabular}
}
\caption{Zero-shot accuracy (\%) on discriminative commonsense QA tasks. Training on human-annotated data causes \textbf{catastrophic forgetting} (avg. $-3.7$ pp), whereas \textbf{CommonSyn} preserves general reasoning capabilities (avg. $-0.1$ pp).}
\label{tab:qa_tasks}
\end{table}

\subsection{Downstream Evaluation}
\label{sec:exp:downstream}
% show synthetic commonsense data improves LLM general reasoning if possible
% on COMVE/ Commonsense QA and other tasks
% Explain what is the purpose of this “knowledge transfer”
% -------------------------------------------

Recall that CommonSyn was created using seed concepts from CommonGen. 
Therefore, it remains a question whether CommonSyn could improve performance in other \ac{GCR} tasks.
To evaluate the zero-shot generalisability of CommonSyn, we fine-tuned Llama-3.1-8B-Instruct on CommonSyn and evaluated it on multiple generative and question answering tasks that require commonsense reasoning.

% whether our synthetic data teaches \textit{generalizable} commonsense reasoning—rather than merely over-fitting to the CommonGen format, we conduct \textbf{zero-shot transfer} experiments on unseen tasks using Llama-3.1-8B-Instruct. 
% We evaluate on both generative and discriminative benchmarks without any further fine-tuning. Task prompts and evaluation protocols are detailed in~\autoref{app:sec:cross_prompt}.

\begin{table}
\centering
\small
\setlength{\tabcolsep}{3.5pt}
\resizebox{\columnwidth}{!}{
\begin{tabular}{l c cc}
\toprule
Method & Unique & \% Unseen & \% Unseen \\
& Concepts & Concepts & Triples \\
\midrule
Human (CommonGen) & 3,677 & 75.36 & 100 \\
\midrule
Direct Generation & 805 & 52.55 & 96.65 \\
Augment (Add-$C$) & 3,791 & 72.80 & 99.65 \\
\textbf{2-Seed Expansion} & \textbf{4,260} & \textbf{75.28} & \textbf{99.62} \\
\bottomrule
\end{tabular}
}
\caption{Comparison of concept set construction strategies ($N=20k$). \textbf{2-Seed Expansion} gets the highest number of unique concepts and maintains high novelty relative to the test set, indicating no data leakage.}
\label{tab:concept_construction}
\end{table}

\begin{figure}[t] 
\centering 
\includegraphics[width=1.0\columnwidth]{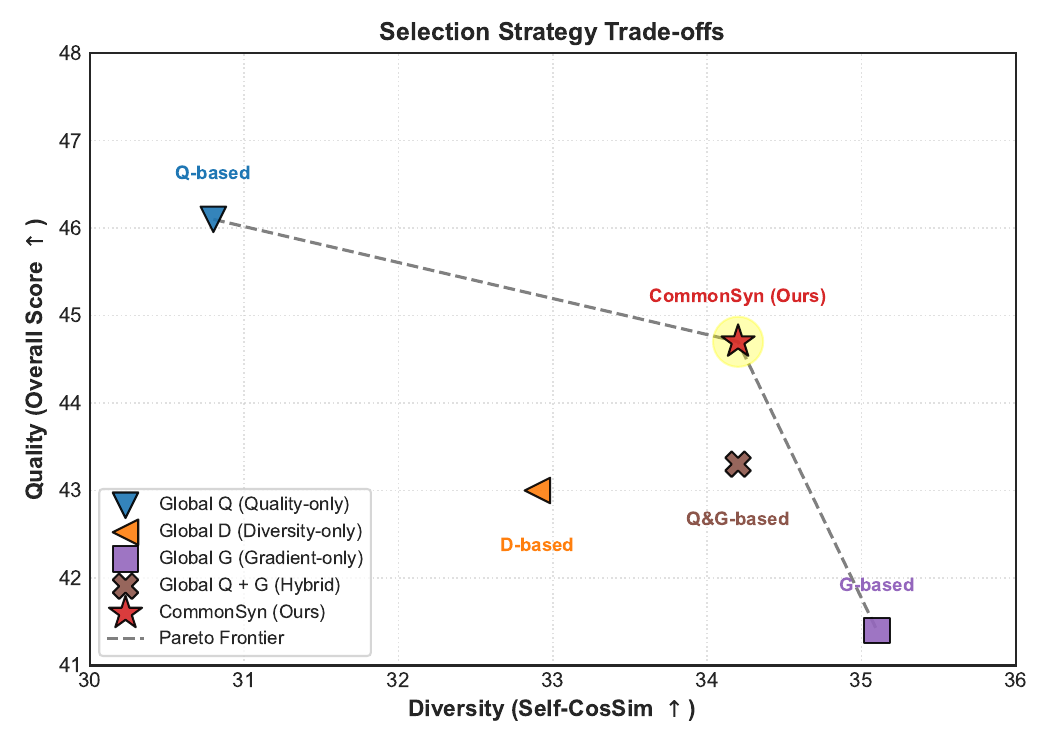} 
\caption{Trade-off between Quality and Diversity. The dashed line indicates the Pareto frontier. \textbf{CommonSyn} (Red Star) strictly outperforms the gradient-hybrid method \textit{Q\&G-based} (Brown Cross) by achieving higher quality at the same diversity level.} 
\label{fig:selection_pareto} 
\end{figure}

\paragraph{Generative Tasks:}
% There are three commonsense reasoning tasks we choose for generative tasks. ComVE~\citep{semeval} aims to generate an explanation given a counterfactual statement for sense-making. 
% $\alpha$-NLG is the task of generating a valid hypothesis about the likely explanations to partially observe pas and future.
% ROCStories~\citep{rocstories} requires a model to write the correct ending to a four-sentence story.
Three \ac{GCR} benchmarks are selected for this evaluation: 
(a) \textbf{ComVE}~\citep{semeval} (generate explanations for counterfactual statements),
(b) \textbf{$\alpha$-NLG}~\citep{alphanlg} (perform abductive reasoning to infer plausible hypotheses), and
(c) \textbf{ROCStories}~\citep{rocstories} (complete a story by generating a coherent ending).
The zero-shot inference prompts are shown in~\autoref{app:sec:cross_prompt}.

Following the CommonGen evaluation protocol, we limit the output length and use GPT-4o to judge the pairwise quality between model generations and human references by the \textbf{Win-Tie} rate.
As shown in~\autoref{tab:gen_transfer} models fine-tuned on \textbf{CommonGen} generalise poorly to other tasks.
Moreover, the quality of the generations drop significantly compared to the vanilla baseline (e.g., the \textbf{Win-Tie} rate nearly halves on ComVE: $75.3 \to 41.7$).
Although diversity metrics appear to be high, we find many hallucinations and commonsense violations as further discussed in~\autoref{app:sec:cross_quan}.

\begin{table}[t]
\centering
\small
\setlength{\tabcolsep}{3.5pt} 
\resizebox{\columnwidth}{!}{
\begin{tabular}{lcccccc}
\toprule
\multirow{2}{*}{\textbf{Strategy}} & \multicolumn{3}{c}{\textbf{Quality} ($\uparrow$)} & \multicolumn{3}{c}{\textbf{Diversity} ($\uparrow$)} \\
\cmidrule(lr){2-4} \cmidrule(l){5-7}
& Win-Tie & Cov. & \textbf{Overall} & S-B4$^\dagger$ & Vendi & S-Cos$^\dagger$ \\
\midrule
Q-based & \textbf{49.0} & 94.1 & \textbf{46.1} & 78.5 & 21.4 & 30.8 \\
D-based & 45.6 & 94.3 & 43.0 & 80.6 & 22.2 & 32.9 \\
G-based & 43.9 & 94.4 & 41.4 & \textbf{83.8} & \textbf{23.0} & \textbf{35.1} \\
Q\&G-based & 45.7 & \textbf{94.7} & 43.3 & 81.8 & 22.7 & 34.2 \\
\midrule
\textbf{CommonSyn} & 47.3 & 94.5 & 44.7 & 82.9 & 22.7 & 34.2 \\
\bottomrule
\end{tabular}
}
\caption{Comparison of selection strategies. \textbf{CommonSyn} outperforms the gradient-based hybrid (\textit{Q\&G-based}) in quality while matching its diversity, validating the effectiveness of our 2-stage selection.}
\label{tab:selection_ablation}
\end{table}

In contrast, models fine-tuned on \textbf{CommonSyn} generalises well to other tasks by consistently outperforming both the vanilla baseline and the CommonGen-finetuned models in terms of quality. 
Furthermore, simultaneous improvements for both quality and diversity can be observed over the vanilla model.
These results suggest that CommonSyn improves the \ac{GCR} capabilities not only for CommonGen, but also for other tasks.

\begin{table*}[t!]
\centering
\small
\setlength{\tabcolsep}{4pt}
\renewcommand{\arraystretch}{1.3} 
\resizebox{0.95\textwidth}{!}{
\begin{tabular}{l p{2.5cm} p{4cm} p{3.5cm} p{4cm}}
\toprule
\textbf{Input Concepts} & \textbf{Human Reference} & \textbf{Vanilla} & \textbf{CommonGen} & \textbf{CommonSyn (Ours)} \\
\midrule
\textit{dog, throw, frisbee, catch} & My dog likes to catch frisbees after I throw them. & As the sun was setting... a energetic dog awaited its owner's signal to throw the frisbee... & A golden colored dog \textbf{throws a frisbee} and catches it. \textcolor{red}{\xmark (Logic)} & I \textbf{throw} the frisbee, and my dog leaps to \textbf{catch} it mid-air. \\
\midrule
\textit{ride, dog, run, shoot, horse} & A dog runs by while the man rides a horse and shoots. & As the rodeo rider rode his horse at full speed, he expertly steered it... (Misses: dog, shoot) & ...shoots a watermelon with a water pistol \textbf{on a dog}. \textcolor{red}{\xmark (Nonsense)} & She \textbf{rode} the \textbf{horse} past a field where her \textbf{dog} \textbf{ran} and someone \textbf{shot} at a deer. \\
\midrule
\textit{instruction, tie, neck, give} & I had to give instructions to tie the tie around her son's neck. & To properly tie a neck tie, follow the instructions carefully... & \textbf{tied up in knots how to} give instructions for tying a tie. \textcolor{red}{\xmark (Grammar)} & In class, the teacher \textbf{gives} detailed \textbf{instruction} on how to properly \textbf{tie} a neck \textbf{tie}. \\
% \midrule
% \textit{top, sunglass, hold, head, mug} & He wore sunglasses on his head while his hands held a top and mug. & ...set the top on the coffee mug... (Misses: sunglass) & An Asian woman is holding a mug with her head up... (Misses: top) & At the cafe, he \textbf{holds} a \textbf{top}-favorite \textbf{mug}, \textbf{sunglasses} resting atop his \textbf{head}. \\
\bottomrule
\end{tabular}
}
\caption{A comparison of sentences generated by different methods. Although \textbf{CommonSyn} generates concise, logically sound, and grammatically correct sentences that fully cover the input concepts, other approaches often suffer from verbosity (Vanilla), commonsense violations (CommonGen), or incorrect grammar.}
\label{tab:case_study}
\end{table*}

\paragraph{Commonsense QA Tasks:}
We further evaluate the generalisability of CommonSyn fine-tuned models on three commonsense QA tasks: CSQA~\citep{csqa}, CSQA2~\citep{csqa2} and PIQA~\citep{piqa}. 
As shown in~\autoref{tab:qa_tasks}, the CommonGen finetuned model exhibits an average accuracy drops of 3.7 percentage points (from 63.8 to 60.1). 

In contrast, the CommonSyn-finetuned model not only avoids this degradation—it even shows slight improvements on CSQA (+0.5 pp) and PIQA (+2.1 pp), resulting in near-identical average performance to the vanilla baseline (63.7 vs. 63.8).

\subsection{Ablation Study}
\label{sec:ablation}
%To study the effect of the different components in our proposal, we perform an ablation study as described next.

\paragraph{Concept Set Construction Strategy} 
\label{sec:exp:ablation:concept}
We compare our \textbf{2-Seed Expansion} strategy against CommonGen and two baselines: \textbf{Direct Generation} (Directly require \ac{LLM} to generate the concept set) and \textbf{Augment} (Add-$C$), which augments an existing concept set first by adding $C$ related concepts and then randomly removing $C$ concepts from it to form a new set.
We sample 20k concept sets for each method and report Unique Concepts and Unseen rates relative to the test set.
As shown in~\autoref{tab:concept_construction}, \textbf{Direct Generation} suffers from \textbf{low diversity} in the generated concept sets, covering only 805 unique concepts as the model regresses to high-frequency generic terms. 
Although \textbf{Augment} improves coverage, it remains bounded by the original distribution. 
In contrast, our \textbf{2-Seed} generates more unique concepts than the human-crafted CommonGen dataset and maintains a high rate of unseen concept composition (unique concept, concept-triples). 
This confirms that two seeds strategy effectively overcomes the coverage of human-written sentences while preventing data leakages.

Furthermore, CommonSyn shows stable performance across different concept set sizes, suggesting high data efficiency.
In~\autoref{tab:concept_scale}, around 20k concept sets are sufficient to saturate performance while remaining cost-effective.
A detailed analysis of dataset scaling is provided in \autoref{sec:exp:ablation:scale}.

\paragraph{Data Selection Strategies} 
\label{sec:exp:ablation:selection}
% Compare the different data selection method % We can compare different selection stratgies selection: gradient-based, mab-based, only on D, only on Q, and find Two-step is more balanced no method dominates the three sources
%---------------------------------------
We compare different data selection methods considering quality, diversity, and gradient and its  hybrid with quality approach.
The number of samples is held constant across comparisons.
As shown in~\autoref{fig:selection_pareto}, these strategies occupy distinct positions in the performance landscape. 
The Quality-based (Q-based) selection strategy, which selects the coreset based on the score given by GPT-4o scorer as mentioned in~\autoref{para:local_select}, maximises quality but suffers from low diversity.
The Gradient-based (G-based) strategy conducts $k$-means clustering using the loss gradient obtained from a proxy model for each sentence and then samples according to cluster sparsity~\citep{jung2025prismatic}.
G-based prioritises diversity over quality.
\textit{Q\&G-based} improves quality but still underperforms CommonSyn. 
Furthermore, gradient-based methods require expensive gradient embeddings (e.g., for Llama-3.1-8B), while CommonSyn achieves a better trade-off at a lower computational cost.
Among all the data selection strategies, CommonSyn shows the optimal balance on the Pareto frontier. 

\subsection{Case Study}
\label{sec:exp:quantitive}
% Show some examples from the Generation
% -------------------------------------------
To show the qualitative differences between models, we compare generations from the vanilla model, CommonGen-finetuned, and our \textbf{CommonSyn} in~\autoref{tab:case_study}.  
We observe that the vanilla model omits input concepts (e.g., missing ``dog'' and ``shoot'' in the second example) and tends to produce verbose text that deviates from the concise style expected in CommonGen. 
The CommonGen-finetuned model arranges the concepts into implausible scenarios (e.g., ``shooting a watermelon with a water pistol on a dog''), violating commonsense constraints.
In contrast, \textbf{CommonSyn} consistently integrates all input concepts into coherent, natural, and logically plausible sentences. 
For example, in the first example, it correctly assigns agency: the human \textit{throws} the frisbee, and the dog \textit{catches} it—aligning both with physical plausibility and task expectations. 
Additional examples are provided in~\autoref{app:tab:full_case_study}.

\section{Conclusion} 
We created CommonSyn, a synthetic dataset to jointly improve both quality and diversity in~\ac{GCR}.
By leveraging a 2-seed expansion strategy and a joint Quality-Diversity selection algorithm, we constructed a dataset that surpasses human-annotated resources in both scale and semantic coverage.
Our results show that models fine-tuned on our dataset exhibit superior Pareto improvements across diverse model families and scales.
CommonSyn also generalises well on unseen tasks. This establishes our framework as a scalable, robust pathway for enhancing the \ac{GCR} capabilities of \acp{LLM}.

\section{Limitations}
Despite the effectiveness of CommonSyn, we acknowledge the following limitations.
First, our dataset is currently restricted to English, a morphologically limited language.
Although extending CommonSyn to multilingual settings is a desirable direction, we are constrained by the scarcity of high-quality human reference benchmarks in other languages.
In particular, CommonGen~\citep{CommonGen}, ComVE~\citep{semeval}, and $\alpha$-NLG~\citep{alphanlg} datasets are specifically designed to evaluate diversified commonsense reasoning only in English.

Furthermore, the quality of synthetic data depends on the quality of the generator. 
Our method relies on Qwen2.5-72B-Instruct for data synthesis and a separate model (Gemini-2.5-Flash) for quality scoring.
Therefore, the final dataset is inherently bounded by the capabilities and biases of these teacher models, and a better generator might lead to higher quality synthetic data, such as by using GPT-4o to generate sentences. However, we used GPT-4o as an evaluator to compare the model outputs and human references. 
Therefore, we cannot directly use it as a generator or quality scorer when creating synthetic data.

We also acknowledge potential intrinsic biases in using GPT-4o as an automated evaluator.
As recent studies indicate, \ac{LLM}-as-a-Judge exhibits systematic vulnerabilities, including verbosity bias and self-enhancement bias~\citep{zheng:2023:judging,shi2025judging,wang:2024large}. 
To mitigate self-enhancement bias, we use an independent \ac{LLM} (GPT-4o) that is distinct from the models being evaluated.
To mitigate verbosity bias, we follow the CommonGen evaluation protocol, which explicitly instructs the evaluator to prefer shorter, more concise sentences.
Given that valid GCR outputs can be open-ended, we rely on Win-Tie rates against human references to ensure robustness. 
Although we selected GPT-4o for its superior performance in NLG evaluations, verifying our findings with other comparable or superior \acp{LLM} remains an important direction for future work.

Our global selection stage requires computing pairwise cosine similarities for the entire candidate pool to determine global diversity scores. 
Although we argue that this is more efficient than gradient-based methods, the computational complexity of pairwise comparisons can still become a bottleneck when scaling to significantly larger candidate pools (e.g., millions of samples).

Our primary evaluation focuses on CommonGen and related \ac{GCR} tasks. 
Although we demonstrate generalisation ability to tasks like ROCStories, our \textbf{2-seed concept expansion} strategy is explicitly designed for \textit{concept-to-text} generation tasks, where the input is a discrete set of keywords.
This module is not directly applicable to standard sequence-to-sequence tasks (e.g., summarisation, translation, or open-ended story generation) where the input typically consists of complete sentences.
Applying our method to such tasks would require redesigning the question generation strategies.

\section{Ethical Consideration}
All experiments conducted in this study use publicly available datasets such as CommonGen, ComVE, and $\alpha$-NLG.
To the best of our knowledge, no personally identifiable information has been included in those datasets and no ethical issues have been reported.

Our study introduces \textit{CommonSyn}, a synthetic dataset for commonsense reasoning generated using \acp{LLM} and it is noteworthy that \acp{LLM} have been reported to encode social biases such as gender or racial biases~\cite{Kaneko:2021,CrowsPairs,Kaneko:2022a}.
Although we evaluated quality and diversity of the generations made by \acp{LLM} and automatic metrics in this work, we defer a systematic evaluation of social or representational biases in the generated content to future work.
We emphasise the importance of evaluating social biases in our synthetic dataset before it is deployed in \ac{NLG} applications used by human users.
We release our dataset with this disclaimer and encourage its users to apply appropriate filtering, debiasing, or mitigation strategies as needed.

\section*{Acknowledgements}
Danushka Bollegala holds concurrent appointments as a Professor at University of Liverpool and as an Amazon Scholar. This paper describes work performed at the University of Liverpool and is not associated with Amazon.

\bibliography{myrefs}

\appendix

\section{Number of Concept Sets} 
\label{sec:exp:ablation:scale}
% Compare the performance under different number of concepts. How the numbers of concept affect the performance? Trade-offs: data size vs Q-D balance
\begin{table}[h]
\centering
\small
\setlength{\tabcolsep}{3.5pt}
\resizebox{\columnwidth}{!}{
\begin{tabular}{c ccc ccc}
\toprule
\multirow{2}{*}{\textbf{Num of Concept Sets}} & \multicolumn{3}{c}{\textbf{Quality} ($\uparrow$)} & \multicolumn{3}{c}{\textbf{Diversity} ($\uparrow$)} \\
\cmidrule(lr){2-4} \cmidrule(l){5-7}
& Win-Tie & Cov. & \textbf{Overall} & S-B4$^\dagger$ & Vendi & S-Cos$^\dagger$ \\
\midrule
5k  & 46.0 & \textbf{94.4} & 43.4 & 82.2 & 22.4 & 33.5 \\
10k & 46.2 & 94.3 & 43.6 & 82.7 & 17.3 & 34.3 \\
15k & 45.0 & \textbf{94.4} & 42.5 & 82.6 & 22.5 & 33.9 \\
\textbf{20k} & \textbf{46.6} & 94.0 & \textbf{43.8} & \textbf{82.8} & \textbf{22.8} & \textbf{34.6} \\
\bottomrule
\end{tabular}
}
\caption{Performance obtained with numbers of unique concept sets (fixed total sentence sample size = 40k). 
Consistently high performance is seen across concept set sizes, indicating high data efficiency of our proposed method.}
\label{tab:concept_scale}
\end{table}

We compare how the diversity of input scenarios (number of unique concept sets) affects performance. 
We fix the total dataset size at 40k sentence samples and vary the number of unique concept sets from 5k to 20k (implying that fewer sets have more sentences per set, and vice versa).

\autoref{tab:concept_scale} shows that the performance is stable across different scales. 
Increasing the unique concept sets from 5k to 20k yields only slight improvement in \textbf{Overall} (quality) and \textbf{self-CosSim} (diversity).
This suggests that \textbf{CommonSyn} captures the core underlying logic of commonsense generation efficiently, without requiring a large possible concept combination.
In addition, $\sim$20k concept sets are sufficient to saturate the model's capability for this task, showing the concept set size in our final dataset as a cost-effective spot.

\section{Case Study for CommonGen Task}
\label{app:sec:full_case_study}

In~\autoref{app:tab:full_case_study}, we compare the multiple generations from vanilla CommonGen and CommonSyn-finetuned model using Qwen2.5-7B-Instruct model. 
For each concept set, we sample four independent generations from each model to illustrate output diversity, concept coverage, and commonsense plausibility.
In the result, we can find although the vanilla model can generate plausible sentences, the sentences it generates are highly redundant. The sentences generated by CommonGen-finetuned model are sometimes commonsense-conflicting (``Dog throws the Frisbee for another dog to catch.''). 
However, the model fine-tuned on our CommonSyn generates both high quality and diversity sentences.

\begin{table*}
\centering
\scriptsize
\setlength{\tabcolsep}{3pt} 
\renewcommand{\arraystretch}{1.1} 
\begin{tabular}{l p{4.0cm} p{3.6cm} p{4.2cm}}
\toprule
\textbf{Input Concepts} & \textbf{Vanilla} & \textbf{CommonGen} & \textbf{CommonSyn (Ours)} \\
\midrule

\textit{take, shot, goal, player} &
\begin{enumerate}[nosep, leftmargin=*, label=\arabic*.] 
    \item The player took a shot with the intention of scoring the goal in the upcoming match.
    \item The player takes a shot with the intention of scoring the goal in the ongoing soccer match.
    \item The goal keeper took a daring shot at goal, hoping to catch the player off guard.
    \item The player takes a shot with the intention of scoring the goal in the ongoing soccer match.
\end{enumerate} &
\begin{enumerate}[nosep, leftmargin=*, label=\arabic*.]
    \item soccer player takes a shot on goal during their match .
    \item football player takes a shot on goal during a friendly match .
    \item soccer player taking a shot for the goal during the match .
    \item football player takes a shot on goal during a training session .
\end{enumerate} &
\begin{enumerate}[nosep, leftmargin=*, label=\arabic*.]
    \item The player takes a shot to score the final goal in the championship.
    \item During the intense match, the player took a bold shot toward the goal.
    \item The soccer player took a powerful shot to score the goal.
    \item The soccer player takes a fierce shot towards the goal, hoping to score.
\end{enumerate} \\
\midrule
\textit{pan, stove, cook, food} &
\begin{enumerate}[nosep, leftmargin=*, label=\arabic*.]
    \item She can pan fry the chicken on the stove to cook the delicious food.
    \item In the morning, I like to cook delicious food on a stove using a pan.
    \item She uses a stove to cook food in a pan.
    \item She uses a pan on the stove to cook food for dinner.
\end{enumerate} &
\begin{enumerate}[nosep, leftmargin=*, label=\arabic*.]
    \item A young boy cooking food with a pan on the stove.
    \item A man is cooking food in a pan on a stove.
    \item A man is cooking food in a pan on a stove.
    \item a man cooks with a frying pan on the stove.
\end{enumerate} &
\begin{enumerate}[nosep, leftmargin=*, label=\arabic*.]
    \item In the morning, he stands at the stove to cook breakfast in a pan.
    \item I use a pan on my stove to cook food every day.
    \item Every Sunday, I cook a hearty meal on my pan over the stove.
    \item He carefully cooked the pasta on the stovetop pan for his family.
\end{enumerate} \\
\midrule
\textit{dog, throw, frisbee, catch} &
\begin{enumerate}[nosep, leftmargin=*, label=\arabic*.]
    \item The dog happily catches the frisbee that was thrown by its owner.
    \item The dog threw the frisbee and then eagerly waited to catch it.
    \item The dog threw the frisbee and then eagerly waited to catch it.
    \item The dog threw the frisbee and then patiently waited to catch it again.
\end{enumerate} &
\begin{enumerate}[nosep, leftmargin=*, label=\arabic*.]
    \item A young boy throwing a frisbee to his dog as he catches it.
    \item A woman is throwing a frisbee for her dog to catch.
    \item Two dogs play tug with a frisbee.
    \item Dog throws the Frisbee for another dog to catch.
\end{enumerate} &
\begin{enumerate}[nosep, leftmargin=*, label=\arabic*.]
    \item My dog stands excitedly as I throw the frisbee; it always loves a good catch.
    \item She throws a frisbee for her dog to catch in the park.
    \item My dog always catches the frisbee I throw during our evening walks.
    \item The owner throws the frisbee for his dog to catch, enjoying the sunny day.
\end{enumerate} \\
\midrule
\textit{coffee, drink, newspaper, read} &
\begin{enumerate}[nosep, leftmargin=*, label=\arabic*.]
    \item He likes to drink coffee while reading the newspaper in the morning.
    \item He likes to read the newspaper while drinking coffee in the morning.
    \item He likes to read the newspaper while drinking coffee in the morning.
    \item He likes to drink coffee while reading the newspaper in the morning.
\end{enumerate} &
\begin{enumerate}[nosep, leftmargin=*, label=\arabic*.]
    \item young woman reading a newspaper while drinking coffee and smoking.
    \item young women drinking a coffee and reading the newspaper at home.
    \item A woman is reading a newspaper while drinking coffee.
    \item The man is drinking coffee while reading a newspaper.
\end{enumerate} &
\begin{enumerate}[nosep, leftmargin=*, label=\arabic*.]
    \item She enjoys a quiet morning with a cup of coffee and the local newspaper, sipping it as she reads.
    \item He likes to drink his coffee while reading the morning newspaper.
    \item Every morning, he drinks a cup of coffee and reads the newspaper.
    \item After work, I sit down to read the newspaper and drink coffee.
\end{enumerate} \\
\midrule

% Row 5
\textit{ball, throw, pitcher, batter} &
\begin{enumerate}[nosep, leftmargin=*, label=\arabic*.]
    \item The pitcher threw the ball to the batter during the baseball game.
    \item The pitcher threw the ball to the batter during the baseball game.
    \item The pitcher threw the ball to the batter during the baseball game.
    \item The pitcher throws the ball to the batter during the baseball game.
\end{enumerate} &
\begin{enumerate}[nosep, leftmargin=*, label=\arabic*.]
    \item The pitcher has just thrown a pitch to home plate and the batter is swinging.
    \item A batter is waiting for the ball to be thrown by the pitcher.
    \item a ball thrown by a pitcher and the batter.
    \item pitcher throws a ball at the batter.
\end{enumerate} &
\begin{enumerate}[nosep, leftmargin=*, label=\arabic*.]
    \item The pitcher throws the ball, but the batter swings and misses, watching it sail past.
    \item The batter waited for the ball to be thrown by the pitcher.
    \item During practice, a young batter focuses on how to react to a thrown ball by the pitcher.
    \item The pitcher throws a curve ball past the batter.
\end{enumerate} \\
\bottomrule
\end{tabular}
\caption{Extended case study showing four sampled generations per model for each concept set. CommonGen often ignores or distorts input concepts, while Vanilla exhibits high redundancy. CommonSyn produces diverse, fluent, and concept-consistent outputs that respect real-world roles and relationships.}
\label{app:tab:full_case_study}
\end{table*}

\section{Downstream Tasks Evaluation}
\label{app:sec:cross_task}
In this section, we provide details of our \textbf{zero-shot} experiments on unseen generative tasks, including qualitative case studies and the inference prompts used in evaluation (see~\autoref{app:sec:cross_prompt} for prompt templates).
There are three commonsense reasoning tasks we choose for generative tasks:
\begin{itemize}
    \item ComVE~\citep{semeval} aims to generate a short explanation given a counterfactual statement to explain why the thing described by statement can not happen in real world. 
    \item $\alpha$-NLG~\cite{alphanlg} is the task of generating a valid hypothesis about the likely explanations to partially observe pas and future. The input is two observations with ordering and the hypothesis generated by the model should design a scenario to connect the observations.
    \item ROCStories~\citep{rocstories} requires a model to write the plausible ending to a four-sentence daily story.
\end{itemize}

\subsection{Case Study on More Tasks}
\label{app:sec:cross_quan}
To better understand the nature of the \textit{quality drop} observed in models fine-tuned on human-annotated CommonGen data, we conduct a qualitative analysis of model outputs across three diverse generative commonsense tasks: \textbf{ComVE}, \textbf{$\alpha$-NLG}, and \textbf{ROCStories}.
~\autoref{tab:cross_task_examples} presents a comparison across three tasks: ComVE, $\alpha$-NLG, and ROCStories.

As shown in~\autoref{tab:cross_task_examples}, although the CommonGen-finetuned model sometimes achieves high lexical diversity (\autoref{tab:gen_transfer}), its generations often suffer from hallucination, logical inconsistency—issues that degrade output quality despite surface-level fluency.

For example, in the \textbf{ComVE} task—which requires explaining why a given statement is counterfactual—the CommonGen model generates: ``The image shows sugar is added to make coffee sour.'' 
This response hallucinates a non-existent visual context, violating the purely textual nature of the task. 
In contrast, our \textbf{CommonSyn}-finetuned model correctly reasons that ``sugar is typically added to make coffee sweeter, not sourer,'' demonstrating grounded commonsense understanding.

Similarly, in \textbf{$\alpha$-NLG}, where the goal is to abductively link two observations, the CommonGen model produces a hypothesis that directly contradicts the second observation (e.g., claiming Timmy \textit{passed} when O2 states he \textit{failed}). The Vanilla model, while coherent, often lacks causal depth. Only \textbf{CommonSyn} consistently generates plausible, fact-consistent explanations that respect both input observations.

These cases show that fine-tuning on human-annotated CommonGen data can induce format overfitting, leading models to prioritise stylistic similar (e.g., descriptive elaboration) over semantic fidelity. 
In contrast, training on \textbf{CommonSyn} shows generalizable reasoning skills and performs robustly on unseen tasks.

\begin{table*}[h]
\centering
\small
\setlength{\tabcolsep}{5pt}
\renewcommand{\arraystretch}{1.4}
\begin{tabular}{p{2.8cm} p{3.8cm} p{3.8cm} p{4.2cm}}
\toprule
\textbf{Task Input} & \textbf{Vanilla (Zero-shot)} & \textbf{CommonGen (Human)} & \textbf{CommonSyn (Ours)} \\
\midrule
\multicolumn{4}{l}{\textit{\textbf{Task 1: ComVE} (Generate explanation for why the statement is false)}} \\
\textbf{Input:} sugar is used to make coffee sour &
That's a counterfactual statement. \newline \textit{(Refusal/Vague)} &
``The image shows sugar is added to make coffee sour.'' \newline \textcolor{red}{\xmark \textit{(Hallucination: no image exists)}} &
Sugar is typically added to make coffee sweeter, not sourer. \newline \textcolor{green}{\cmark \textit{(Correct commonsense reasoning)}} \\

\textbf{Input:} The inverter was able to power the continent. &
It defies the fundamental laws of physics and engineering. &
The inverter in the car wasn't able to power the entire continent of Europe. &
The inverter, no matter how powerful, cannot power an entire continent due to its size and energy requirements. \\
\midrule
\multicolumn{4}{l}{\textit{\textbf{Task 2: $\alpha$-NLG} (Generate a hypothesis connecting Observation 1 and 2)}} \\
\textbf{O1:} Timmy was running fast to class. \newline \textbf{O2:} Timmy later found out that he failed. &
Timmy likely ran to class because he forgot that the exam was scheduled for a specific time and he didn't want to be late. &
Because Timmy was running fast to class, he made it to his exam on time and therefore he passed. \newline \textcolor{red}{\xmark \textit{(Contradicts O2: he failed)}} &
Timmy's need to be on time for the exam, combined with his earlier desire to be punctual, suggests he likely rushed to class, possibly missing key information. \newline \textcolor{green}{\cmark \textit{(Plausible inference consistent with failure)}} \\

\textbf{O1:} Music was Kat's life. \newline \textbf{O2:} She completed her studies with honors and is now a renowned pianist. &
Kat enjoyed it so much that she played all of the time. &
Music was Kat's life and she is now a renowned pianist. \newline \textcolor{red}{\xmark \textit{(Mere repetition, no causal link)}} &
Kat's dedication to music fueled her academic success, leading to her distinguished career as a pianist. \\
\midrule
\multicolumn{4}{l}{\textit{\textbf{Task 3: ROCStories} (Generate a coherent story ending)}} \\
\textbf{Context:} Harriet was reading a great mystery book. She read it all evening and quietly closed the book. &
As she set the book on the nightstand, her phone buzzed with an incoming text from her best friend, Emily. \newline \textit{(Introduces irrelevant new element)} &
The cover featured a silhouette of a lone person in a dense forest with a cabin to the right. \newline \textcolor{red}{\xmark \textit{(Describes cover, not ending!)}} &
Feeling the tension dissipate, Harriet leaned back into the couch, her eyes already drifting shut. \newline \textcolor{green}{\cmark \textit{(Natural, thematically consistent closure)}} \\
\bottomrule
\end{tabular}
\caption{Qualitative comparison on downstream generative tasks. Models fine-tuned on human-authored CommonGen data generalise poorly: they hallucinate non-existent images (ComVE), contradict input facts ($\alpha$-NLG), or generate off-topic descriptions (ROCStories). In contrast, \textbf{CommonSyn} shows robust commonsense reasoning and task adherence across unseen domains.}
\label{tab:cross_task_examples}
\end{table*}

\subsection{Downstream Tasks Prompting}
\label{app:sec:cross_prompt}

In this section, we show the generation prompts and evaluation prompts in these downstream tasks.
To limit the surface diversity of sentence groups due to their length, we set the generation length to 22 for each generative task (\autoref{tab:inference_prompts}).

In~\autoref{fig:eval_prompts}, we present the evaluation prompts from each downstream generative task. 
Apart from the plausibility, the generation also needs to meet the requirements for different tasks. 
When evaluate the generation for each task, we also randomly shuffle the model generated sentences and the human-written references so that the ordering will not affect the \ac{LLM} evaluation result.

% Generation in a table and evaluation in three small figures.
\begin{table*}[h]
\centering
\small
\begin{tabular}{p{3cm} p{10cm}}
\toprule
\textbf{Task} & \textbf{Inference Prompt} \\
\midrule
ComVE & Given an implausible or counterfactual statement, generate one short explanation ($\leq 22$ words) that explains why it is implausible or counterfactual using background commonsense knowledge. \\
\midrule
$\alpha$-NLG & Given an initial observation and a later observation, generate a short hypothesis ($\leq 22$ words) that bridges Observation 1 and Observation 2 using background commonsense knowledge. \\
\midrule
ROCStories & Read the following 4-sentence story context and write a short and plausible ending ($\leq 22$ words) to this story using background commonsense knowledge. \\
\bottomrule
\end{tabular}
\caption{Inference prompts used for zero-shot evaluation on downstream generative tasks. All outputs are constrained to $\leq 22$ words.}
\label{tab:inference_prompts}
\end{table*}

\begin{figure*}[h]
\centering
\small
\begin{minipage}{0.95\textwidth}
\setlength{\parskip}{6pt}
\textbf{ComVE Evaluation Prompt:} \\
\fbox{\parbox{\dimexpr\linewidth-2\fboxsep-2\fboxrule\relax}{
Given an implausible or counterfactual statement, we ask models to generate a short explanation of why the statement is implausible or counterfactual. \\
Statement: \{\$input\} \\
Model A: \{\$candidate\_A\} \\
Model B: \{\$candidate\_B\} \\[4pt]
Your Task: Choose the better explanation. Decide which model’s output better explains why the statement is implausible. \\
Rules: \\
– A good explanation should point out the everyday commonsense reason why the statement is unrealistic. \\
– Prefer simple, intuitive explanations that a person would naturally give. \\
– Prefer explanations that highlight the obvious mismatch with real-world scenes. \\
– If both explanations are equally good or equally flawed, choose "tie". \\
Now, please output your choice ("A", "B", or "tie").
}}

\vspace{8pt}

\textbf{$\alpha$-NLG Evaluation Prompt:} \\
\fbox{\parbox{\dimexpr\linewidth-2\fboxsep-2\fboxrule\relax}{
Given two observations (O1 and O2) that describe the beginning and end of a short scenario, we ask models to generate a "Hypothesis" (a middle sentence) that explains what happened in between to cause the transition from O1 to O2. \\
Input Data: "\{\$input\}" \\
Model A: "\{\$candidate\_A\}" \\
Model B: "\{\$candidate\_B\}" \\[4pt]
Your Task: Choose the better hypothesis. Decide which model's output creates a more plausible and coherent story bridge between the two observations. \\
Rules: \\
– A good abductive explanation should provide a plausible cause or hidden event that makes the transition from Observation 1 to Observation 2 reasonable. \\
– Prefer explanations that reflect everyday commonsense and real-world causal relations. \\
– Prefer explanations that clearly “bridge the gap” between the two observations. \\
– Avoid explanations that contradict either observation or introduce unlikely events. \\
– If both explanations are equally good or equally flawed, choose "tie". \\
Now, please output your choice ("A", "B", or "tie").
}}

\vspace{8pt}

\textbf{ROCStories Evaluation Prompt:} \\
\fbox{\parbox{\dimexpr\linewidth-2\fboxsep-2\fboxrule\relax}{
Given the beginning of a short story, we ask models to generate a plausible and coherent continuation. \\
Story Prompt: "\{\$input\}" \\
Model A Ending: "\{\$candidate\_A\}" \\
Model B Ending: "\{\$candidate\_B\}" \\[4pt]
Your Task: Choose which ending provides a more coherent and plausible completion of the story based on the criteria. \\
Evaluation Criteria: \\
– Relevance: Incorporate relevant details from the prompt. \\
– Coherence: Events follow clear causal and temporal progression. \\
– Clarity: Easily understandable with proper grammar. \\
– Commonsense Knowledge: Demonstrate correct real-world knowledge. \\
– Creativity: Provide a fresh and interesting perspective. \\
– Length: Adhere to specified length requirements. \\
If both continuations are equally good or equally flawed, choose “tie”. \\
Now, output your choice ("A", "B", or "tie").
}}
\end{minipage}
\caption{Evaluation prompts used by GPT-4o to judge model generations in pairwise comparison. Each prompt defines task-specific criteria for selecting the better output between the model output and Human reference.}
\label{fig:eval_prompts}
\end{figure*}

\section{Generation Prompting}
\label{app:sec:gen_prompt}
In this section, we detail the prompting strategies used to construct our synthetic dataset CommonSyn. 
First, we expand each 2-seed concept pair into a coherent event-centric concept set using the instruction in~\autoref{fig:synthetic_prompt}. 
Then, for each expanded concept set, we generate candidate sentences via three distinct prompting paradigms—Dynamic Few-Shot, Multi-sentence Few-Shot, and Chain-of-Thought as specified in~\autoref{fig:sentence_gen_prompts}. 
All generated sentences are constrained to $\le 22$ words in the prompts.

\begin{figure*}[t] 
\centering
\small
\begin{minipage}{0.95\textwidth}
\setlength{\parskip}{6pt}

\textbf{Synthetic Data Generation Prompt (2-Seed Concept Expansion):} \\
\fbox{\parbox{\dimexpr\linewidth-2\fboxsep-2\fboxrule\relax}{
Instruction: You are an expert in commonsense knowledge. \\
Given a seed of two keywords, complete the set by adding EXACTLY \#\#num\_to\_add\#\# keywords.

Your goal is to produce a concept set that can naturally form ONE plausible daily-life event or scene, \\
which can be expressed in ONE single sentence under 22 words.

Rules for ADDED keywords:
\begin{itemize}
    \item Each keyword must be a common noun or verb in dictionary form.
    \item DO NOT use prepositions, articles, or pronouns.
    \item ADDED keywords must form a coherent \textit{action chain} with the seed keywords.
    \item The combined keywords (seed + added) MUST be usable to construct one daily-life scene or action.
    \item Avoid overly abstract or static concepts.
    \item Output ONLY the added keywords, separated by commas.
    \item You MUST NOT output any keyword that appears in the seed, even if it seems logically appropriate.
\end{itemize}

Here are some examples of varied lengths:

---\\
Example 1: \\
Seed: passenger, train \\
Output: station, run \\
---\\
...

Your Task:\\
Seed: \#seed\_keywords\#\\
Output:\\
}}
\end{minipage}
\caption{Prompt template used to expand 2-seed concept sets during synthetic data generation. This instruction guides an LLM to add contextually relevant keywords that enable the construction of a single coherent, everyday scenario.}
\label{fig:synthetic_prompt}
\end{figure*}

% 3 sentence generation method
\begin{figure*}[h]
\centering
\small
\begin{minipage}{0.95\textwidth}
\setlength{\parskip}{6pt}

\textbf{Prompt for Few-Shot and Multi-Sentence Generation ($D_{\text{dyn}}$, $D_{\text{ms}}$):} \\
\fbox{\parbox{\dimexpr\linewidth-2\fboxsep-2\fboxrule\relax}{
You generate fluent and commonsense-bearing English sentences under strict formatting constraints. Do not add explanations or numbering. Output only the requested sentences.

\vspace{4pt}
Instruction: Given \#\#num\_to\_add\#\# keywords, generate exactly \#\#N\#\# commonsense-bearing and diverse English sentences. \\
Each sentence MUST contain ALL the required keywords (inflectional variants are allowed, e.g., stand→stood/stands).

Requirements:
\begin{itemize}
    \item The sentence must describe a logically possible everyday situation.
    \item It must contain \textbf{ALL} the provided keywords (inflectional variants allowed).
    \item Keep the sentence concise ($\leq 22$ words).
    \item Separate sentences with a single TAB character (\textbackslash t).
    \item Do not add explanations, numbering, or commentary.
    \item Output exactly \#\#N\#\# sentence(s).
\end{itemize}

Diversity rule:
\begin{itemize}
    \item Vary subject, perspective (first/third person), tone, or setting to ensure distinct expressions.
    \item Use natural variation while maintaining plausibility.
\end{itemize}

Keywords: \#\#problem\#\# \\
Use the few-shot examples below as inspiration for style and structure, \textbf{not as templates to copy}.
}}

\vspace{10pt}

\textbf{Prompt for Chain-of-Thought Guided Generation ($D_{\text{cot}}$):} \\
\fbox{\parbox{\dimexpr\linewidth-2\fboxsep-2\fboxrule\relax}{
You must follow formatting constraints. Do not add numbering or commentary outside the required reasoning and final sentence. Output both a reasoning step and the final sentence for each example.

\vspace{4pt}
Instruction: Given \#\#num\_to\_add\#\# keywords, generate exactly ONE commonsense-bearing English sentence based on your lifestyle and background. \\
Each sentence MUST contain ALL the required keywords (inflectional variants are allowed, e.g., stand→stood/stands).

Requirements:
\begin{enumerate}
    \item First write a reasoning or description to explain the underlying commonsense connection of the keywords. Start with \textbf{``Let's think step by step:''}
    \item Then, on the next line, output ONE realistic English sentence that contains ALL the keywords.
    \item Keep the sentence concise ($\leq 22$ words).
\end{enumerate}

Formatting constraints:
\begin{itemize}
    \item Use exactly \textbf{one reasoning paragraph ($\ge 4$ sentences)} and \textbf{one sentence} per generation.
    \item Separate each generation with a blank line.
    \item Do not add numbering, commentary, or bullet points.
\end{itemize}

Keywords: \#\#problem\#\#
}}

\end{minipage}
\caption{Prompts used to generate synthetic sentences from expanded concept sets. The top prompt is shared by both dynamic few-shot ($D_{\text{dyn}}$) and multi-sentence few-shot ($D_{\text{ms}}$) strategies; the bottom prompt enforces explicit reasoning via chain-of-thought ($D_{\text{cot}}$). All outputs are constrained to $\leq 22$ words and filtered for full keyword coverage.}
\label{fig:sentence_gen_prompts}
\end{figure*}

\section{Evaluation Prompting}
\label{app:sec:eval_prompt}
In this section, we present the evaluation prompts used to assess the quality of sentences. 
The prompt used to evaluate sentence plausibility per concept set and return score for sentences is shown in~\autoref{fig:prompt:quality_scorer_detailed}.

In~\autoref{fig:commongen_eval_prompt}, the prompt is used to assess the quality of model-generated outputs against human-written references from the CommonGen test set~\citep{CommonGen}. 
Following the official evaluation protocol from the \href{https://github.com/allenai/CommonGen-Eval}{CommonGen-Eval} repository, we employ pairwise comparison prompts that instruct GPT-4o to select the better sentence based on naturalness, concept coverage, and plausibility of everyday scenarios.

To evaluating the human alignment of the evaluating prompts, we have conducted a human evaluation on a randomly sampled subset of 50 examples.
Specifically, we asked humans to annotate the samples with the same instruction as used by the LLM quality evaluator. The human judgments showed 81.6\% agreement and 0.648 Cohen's Kappa with GPT-4o rankings on average, showing a substantial level of agreement, confirming the reliability of our LLM-based evaluation. 
\begin{figure*}[h]
\centering
\small
\begin{minipage}{0.95\textwidth}
\setlength{\parskip}{6pt}

\textbf{Quality Scoring Prompt for Candidate Sentences (Per Concept Set):} \\
\fbox{\parbox{\dimexpr\linewidth-2\fboxsep-2\fboxrule\relax}{
You are an expert evaluator of sentence quality and commonsense reasoning.

I will give you a set of concepts, and \{num\_sentences\} candidate sentences generated using different prompting strategies.

Your task:
\begin{enumerate}
    \item Score each sentence independently from 1 to 10.
    \item Higher score = higher quality and commonsense correctness.
    \item All scores must be integers.
    \item Use the full range (1–10) with these guidelines:
    \begin{itemize}
        \item \textbf{1–3 (poor):} Incorrect, implausible, ungrammatical, or fails to use the concepts meaningfully.
        \item \textbf{4–6 (average):} Mostly correct but may have minor issues in clarity, grammar, or concept integration.
        \item \textbf{7–8 (good):} Clear, fluent, plausible sentences that use the concepts well.
        \item \textbf{9–10 (excellent):} Exceptional clarity, realism, and full concept integration. No errors.
    \end{itemize}
    \item If a sentence is marked as ``[EMPTY]'', assign it a score of 1.
\end{enumerate}

Evaluate based on:
\begin{itemize}
    \item \textbf{Commonsense correctness}: Is the event plausible and realistic?
    \item \textbf{Concept coverage}: Are the concepts used meaningfully together?
    \item \textbf{Clarity and grammar}: Is the sentence well-formed?
\end{itemize}

Concept set: \\
\{concept\_set\}

Candidate sentences: \\
\{sentence\_list\}
}}

\end{minipage}
\caption{Prompt used by the quality scorer (Gemini-2.5-flash) to assign plausibility scores $Q(y) \in [1,10]$ to each candidate sentence given a concept set $\mathcal{C}$. Scores guide filtering ($Q(y) < 4$ discarded) and subsequent local diversity selection.}
\label{fig:prompt:quality_scorer_detailed}
\end{figure*}

\begin{figure*}[h]
\centering
\small
\begin{minipage}{0.95\textwidth}
\setlength{\parskip}{6pt}

\textbf{CommonGen Evaluation Prompt (Main Task):} \\
\fbox{\parbox{\dimexpr\linewidth-2\fboxsep-2\fboxrule\relax}{
Given several concepts (i.e., nouns or verbs), we ask models to write a short and simple sentence that contains all the required words. \\
The sentence should describe a common scene in daily life, and the concepts should be used in a natural way.

\vspace{4pt}
Concepts: \{\$input\} \\
Model A: \{\$candidate\_A\} \\
Model B: \{\$candidate\_B\}

\vspace{4pt}
Your Task: Your task is to choose a better sentence from the two candidates. Decide which model's sentence is better in terms of the naturalness and commonness of the scenes they describe.

\vspace{4pt}
Rules:
\begin{itemize}
    \item A better sentence should describe a common scene in daily life, and all concepts should be used in a natural way.
    \item You should prefer sentences that use all given concepts.
    \item A simpler and shorter sentence is preferred if it describes the same scene as the other sentence.
    \item If you think both sentences are equally good or bad, please choose "tie".
\end{itemize}

Now, please output your choice ("A" or "B" or "tie").
}}

\end{minipage}
\caption{Evaluation prompt used to compare model-generated sentences against human references on the CommonGen task. This prompt—adapted from the official \texttt{CommonGen-Eval} repository—guides GPT-4o to judge based on naturalness, concept coverage.}
\label{fig:commongen_eval_prompt}
\end{figure*}

\section{Full Result Table}
\label{app:sec:full_table}
~\autoref{tab:full_results} shows the full result table for all the models we evaluate our CommonSyn synthetic data.
We do experiments on 11 models and the model fine-tuned on CommonSyn consistently outperforms the vanilla models and CommonGen-finetuned models.
\begin{table*}[h!]
\centering
\resizebox{\textwidth}{!}{%
\begin{tabular}{ll ccc c cccc}
\toprule
\multirow{2}{*}{\textbf{Model}} & \multirow{2}{*}{\textbf{Dataset}} & \multicolumn{3}{c}{\textbf{Quality Metrics} ($\uparrow$)} & & \multicolumn{4}{c}{\textbf{Diversity Metrics} ($\uparrow$)} \\
\cmidrule{3-5} \cmidrule{7-10}
 & & Win-Tie & Cov. & Overall & & S-BLEU3 & S-BLEU4 & Vendi & S-Cos \\
\midrule
\multirow{3}{*}{Llama-3.1-8B-Inst} 
 & Vanilla & 19.0 & 84.7 & 16.1 & & 73.4 & 80.3 & \textbf{22.8} & \textbf{35.1} \\
 & CommonGen & 31.7 & 85.8 & 27.2 & & \textbf{76.4} & \textbf{84.70} & 21.2 & 30.7 \\
 & \textbf{CommonSyn} & \textbf{47.3} & \textbf{94.5} & \textbf{44.7} & & 75.5 & 82.9 & 22.7 & 34.2 \\
\midrule
\multirow{3}{*}{Llama-3.2-1B-Inst} 
 & Vanilla & 6.0 & 64.3 & 3.86 & & \textbf{79.5} & 84.8 & \textbf{27.2} & \textbf{49.7} \\
 & CommonGen & 20.9 & 84.0 & 17.56 & & 79.2 & \textbf{87.0} & 21.2 & 34.1 \\
 & \textbf{CommonSyn} & \textbf{31.1} & \textbf{89.6} & \textbf{27.87} & & 79.2 & 86.1 & 24.2 & 38.3 \\
\midrule
\multirow{3}{*}{Llama-3.2-3B-Inst} 
 & Vanilla & 7.8 & 65.5 & 5.11 & & 72.8 & 78.8 & \textbf{24.8} & \textbf{41.2} \\
 & CommonGen & 26.3 & 86.9 & 22.85 & & \textbf{77.0} & \textbf{84.8} & 21.8 & 32.2 \\
 & \textbf{CommonSyn} & \textbf{42.0} & \textbf{92.4} & \textbf{38.81} & & 76.4 & 83.4 & 23.1 & 35.4 \\
\midrule
\multirow{3}{*}{Qwen-2.5-1.5B-Inst} 
 & Vanilla & 23.0 & 79.6 & 18.31 & & 77.7 & 84.9 & 22.1 & 33.3 \\
 & CommonGen & 20.6 & 81.1 & 16.71 & & \textbf{80.0} & \textbf{87.5} & 23.8 & 37.8 \\
 & \textbf{CommonSyn} & \textbf{29.3} & \textbf{84.0} & \textbf{24.61} & & 79.6 & 86.7 & \textbf{24.1} & \textbf{38.2} \\
\midrule
\multirow{3}{*}{Qwen-2.5-3B-Inst} 
 & Vanilla & 19.1 & 87.9 & 16.79 & & 70.7 & 77.7 & 21.3 & 30.6 \\
 & CommonGen & 27.6 & 87.1 & 24.04 & & \textbf{75.7} & \textbf{84.1} & 21.9 & 32.4 \\
 & \textbf{CommonSyn} & \textbf{40.3} & \textbf{91.6} & \textbf{36.91} & & 75.6 & 83.0 & \textbf{22.8} & \textbf{34.5} \\
\midrule
\multirow{3}{*}{Qwen-2.5-7B-Inst} 
 & Vanilla & 42.9 & 92.3 & 39.60 & & 39.2 & 44.1 & 15.8 & 16.3 \\
 & CommonGen & 32.5 & 88.6 & 28.80 & & 73.2 & \textbf{81.8} & 21.4 & 31.2 \\
 & \textbf{CommonSyn} & \textbf{48.4} & \textbf{93.0} & \textbf{45.01} & & \textbf{73.7} & 81.0 & \textbf{22.5} & \textbf{33.7} \\
\midrule
\multirow{3}{*}{Qwen-2.5-14B-Inst} 
 & Vanilla & 45.2 & 95.9 & 43.35 & & 42.4 & 47.5 & 16.5 & 18.6 \\
 & CommonGen & 38.5 & 92.9 & 35.77 & & 70.8 & 79.6 & 20.9 & 29.7 \\
 & \textbf{CommonSyn} & \textbf{52.0} & \textbf{96.0} & \textbf{49.92} & & \textbf{73.2} & \textbf{80.6} & \textbf{22.1} & \textbf{32.6} \\
\midrule
\multirow{3}{*}{Gemma-3-1B-IT} 
 & Vanilla & 18.2 & 83.9 & 15.27 & & 45.5 & 51.1 & 17.2 & 21.0 \\
 & CommonGen & 16.5 & 84.9 & 14.01 & & 79.8 & \textbf{87.6} & 22.0 & 32.6 \\
 & \textbf{CommonSyn} & \textbf{23.3} & \textbf{91.8} & \textbf{21.39} & & 78.7 & 85.7 & \textbf{23.6} & \textbf{36.6} \\
\midrule
\multirow{3}{*}{Gemma-3-4B-IT} 
 & Vanilla & 28.7 & 88.4 & 25.34 & & 37.7 & 42.3 & 16.1 & 17.4 \\
 & CommonGen & 28.7 & 91.0 & 26.12 & & 74.9 & \textbf{83.2} & 21.2 & 30.6 \\
 & \textbf{CommonSyn} & \textbf{41.2} & \textbf{95.5} & \textbf{39.35} & & \textbf{75.4} & 82.6 & \textbf{22.4} & \textbf{33.4} \\
\midrule
\multirow{3}{*}{Gemma-3-12B-IT} 
 & Vanilla & 43.2 & 94.0 & 40.63 & & 31.7 & 35.7 & 14.7 & 13.6 \\
 & CommonGen & 35.9 & 94.1 & 33.78 & & 72.2 & 80.4 & 20.5 & 28.7 \\
 & \textbf{CommonSyn} & \textbf{51.1} & \textbf{97.4} & \textbf{49.77} & & \textbf{73.7} & \textbf{81.2} & \textbf{21.8} & \textbf{31.7} \\
\midrule
\multirow{3}{*}{Gemma-3-27B-IT} 
 & Vanilla & 31.9 & 86.7 & 27.66 & & 27.0 & 30.2 & 15.0 & 14.7 \\
 & CommonGen & 40.3 & 95.6 & 38.53 & & 70.7 & 79.2 & 20.3 & 28.1 \\
 & \textbf{CommonSyn} & \textbf{53.6} & \textbf{97.9} & \textbf{52.47} & & \textbf{71.7} & \textbf{79.2} & \textbf{21.4} & \textbf{30.7} \\
\bottomrule
\end{tabular}%
}
\caption{Full experimental results across all 11 evaluated models. For S-BLEU and S-Cos, higher scores indicate greater diversity.}
\label{tab:full_results}
\end{table*}
\end{document}